\documentclass{article}

\usepackage[preprint]{neurips_2026}
\makeatletter
\def\@noticestring{}
\makeatother

\usepackage[utf8]{inputenc}
\usepackage[T1]{fontenc}
\usepackage{hyperref}
\usepackage{url}
\usepackage{booktabs}
\usepackage{amsfonts}
\usepackage{nicefrac}
\usepackage{microtype}
\usepackage{xcolor}
\usepackage{overpic}

\usepackage{algorithm}
\usepackage{algpseudocode}

\usepackage{graphicx}
\usepackage{wrapfig}
\usepackage{amsmath}
\usepackage{amssymb}
\usepackage{mathtools}
\usepackage{amsthm}
\usepackage{array}
\usepackage{fancybox}
\usepackage{tabularx}
\usepackage{subcaption}
\usepackage[most]{tcolorbox}
\usepackage{tikz}
\usepackage[textsize=tiny]{todonotes}
\usepackage[capitalize,noabbrev]{cleveref}

\definecolor{myLightBlue}{RGB}{230,242,255}

\renewcommand{\eqref}[1]{(\ref{#1})}
\theoremstyle{plain}
\newtheorem{theorem}{Theorem}

\theoremstyle{definition}
\newtheorem{definition}{Definition}

\theoremstyle{remark}

\title{Geometric Analysis of Token Selection in Multi-Head Attention}

\author{%
  Timur Mudarisov
  \And
  Mikhal Burtsev
  \And
  Tatiana Petrova
  \And
  Radu State
}

\begin{document}

\maketitle

\begin{abstract}
We present a geometric framework for analysing multi-head attention in large language models. Without modifying the attention mechanism, we view standard attention through a top-\(N\) selection lens and study attention-scaled value contributions directly in value-state space. We define geometric precision, recall, and an extremal separability score, and derive sink-aware deterministic bounds controlled by selected--unselected inversions. Empirically, across LLaMA, Gemma, and Mistral models, top-attention subsets are more separable than random subsets, while the sink token has a distinct norm and direction that materially changes the selected geometry. We further introduce a parameter-free source-winner taxonomy of attention heads into \emph{Retriever}, \emph{Mixer}, and \emph{Reset} regimes. Under equalized single-document preprocessing, Retriever heads are consistently the minority (\(5.8\)--\(17.2\%\)), whereas the dominant regime alternates between Mixer and Reset across architectures but remains stable between OpenWebText and WikiText. The regimes exhibit distinct mechanistic signatures and provide a useful, though model-dependent, signal for head sparsification. A critical first-layer Gemma head explains the otherwise extreme sparsification outlier and is analysed separately.
\end{abstract}

\section{Introduction}

The attention mechanism is a context-dependent weighting operation that allows a model to aggregate information from its input history~\cite{bahdanau2016neuralmachinetranslationjointly}. 
In the Transformer architecture~\cite{vaswani2023attentionneed}, stacked self-attention and feed-forward layers replaced recurrence and convolutions as the dominant backbone for sequence modelling, and now form the core of modern large language models (LLMs).

A large theoretical literature studies what attention can compute. 
Transformers with suitable positional encodings are universal approximators and can express rich sequence-to-sequence functions~\cite{yun2020transformersuniversalapproximatorssequencetosequence,perez2021turing,bhattamishra2020ability,mudarisov2025limitationsnormalizationattentionmechanism}. 
Another line of work interprets attention as a normalized kernel smoother, motivating efficient linear and kernelized variants with controlled approximation error~\cite{choromanski2022rethinkingattentionperformers}. 
These perspectives clarify expressivity and approximation, but they do not directly describe the geometry of the tokens selected by a trained attention head.

Geometry appears in attention in two different ways. 
First, attention weights lie on the probability simplex: softmax can be viewed as an entropy-regularized optimization, while sparse alternatives such as sparsemax and entmax induce structured low-dimensional solutions~\cite{martins2016softmaxsparsemaxsparsemodel,peters2019sparsesequencetosequencemodels}. 
Second, geometry shapes the representation space itself: rotary positional embeddings act through rotations that encode relative phase~\cite{su2023roformerenhancedtransformerrotary}, and hyperbolic representation spaces have been used to model hierarchical structure~\cite{ganea2018hyperbolicentailmentconeslearning}. 
In this work, we focus on a third geometric object: the attention-scaled value contributions \(\alpha_i v_i\) produced by a fixed layer and head.

Recent empirical work has shown that attention contains strong extreme-token phenomena. 
The attention sink is a robust effect in which early tokens, often the first token, receive disproportionate attention mass~\cite{xiao2023efficient,gu2025attentionsinkemergeslanguage}. 
Other studies show that heads specialize into active and dormant roles~\cite{guo2024activedormantattentionheadsmechanistically}, and token-selection methods such as OrthoRank exploit geometry relative to the sink direction~\cite{shin2025orthoranktokenselectionsink}. 
These results suggest that attention is not just diffuse averaging: it contains structured selection and head-level specialization. 
However, we still lack a simple framework that asks whether the tokens selected by attention form a geometrically separable class in value-state space.

%\paragraph{Main idea: attention as a geometric classifier.}
We view a single attention head as inducing a classifier over attention-scaled value contributions. 
For a fixed query position, let \(I_N\) be the set of top-\(N\) tokens by attention weight and define
\[
    y_i=\alpha_i v_i,
    \qquad
    s_N=\sum_{i\in I_N} y_i .
\]
The question is not whether \(I_N\) is semantically correct, but whether the selected contributions \(y_i\), \(i\in I_N\), are geometrically separated from the remaining contributions around the aggregate \(s_N\). 
We measure this using geometric precision and recall.
%: precision asks whether the neighborhood of \(s_N\) is pure, while recall asks whether selected tokens remain close to their own aggregate. 
This turns top-attention selection into a diagnostic classification problem in value-state space.

%\paragraph{Theory.}
Our theoretical analysis is based on selected-unselected inversions. 
An inversion occurs when an unselected contribution is at least as close to \(s_N\) as a selected contribution. 
We show that geometric precision and recall are controlled by the number of such inversions. 
We then specialize this result to a sink-aware constant-norm regime motivated by empirical observations: non-sink value states have nearly constant norm within a layer and head, while the sink token has a smaller scale and a signed directional relationship with non-sink tokens. 
Under these assumptions, only a subset of selected-unselected pairs can become inversions; the resulting count \(K_N^{\mathrm{sink}}\) gives deterministic lower bounds for precision and recall. 
The bounds are intentionally conservative: they certify separability from worst-case coherence rather than tightly predicting empirical metrics.

%\paragraph{Empirical picture.}
We evaluate the framework on LLaMA, Gemma, and Mistral models using WikiText-103 and OpenWebText. 
First, top-attention subsets are consistently more geometrically separable than random subsets of the same size, especially for sparse and intermediate \(N\). 
Second, the empirical value-state geometry supports the assumptions used in the theorem: non-sink value norms are concentrated within layer-head pairs, while the sink token has a distinct norm and is often negatively aligned with non-sink directions. 
Third, removing the sink from top-\(N\) selection substantially changes the separability score, showing that the sink is an active geometric component rather than a background artifact. 
Finally, a parameter-free source-winner rule induces a taxonomy of \emph{Retriever}, \emph{Mixer}, and \emph{Reset} regimes. The assignments are stable across the two corpora under equalized preprocessing, while their prevalence and their effect on sparsification remain architecture-dependent.

Our contributions:

  \textbf{1. Geometric classifier view of attention.}
    We recast top-\(N\) attention as a geometric classification problem in
    value-state space and introduce precision, recall, and an extremal
    separability score $F_N$.% for attention-scaled value contributions.
    
\textbf{2. Sink-aware bounds on geometric separability.}
    We derive sink-aware lower bounds on $F_N$ under empirically supported
    assumptions on the norm and directional structure of value contributions,
    explicitly accounting for the sink token's anomalous geometry.
    
\textbf{3. Empirical validation across architectures.}
    Across LLaMA, Gemma, and Mistral models, we show that top-attention sets are more separable than random sets, that value norms and sink cosines support the assumptions, and that sink-aware bounds act as conservative certificates of separability.
    
\textbf{4. A parameter-free geometric head taxonomy.}
    We classify every head from the plurality of its contribution-maximizing
    source positions, with no thresholds, margins, or fitted parameters.
    Retriever heads are current-token dominated, Mixer heads are sink-mediated,
    and Reset heads have no privileged source. Retriever is the minority regime
    in every model (\(5.8\)--\(17.2\%\)); the dominant regime is architecture
    dependent, while OpenWebText--WikiText shifts remain below five percentage
    points.

\textbf{5. Functional validation and limitations via head sparsification.}
    The taxonomy provides a useful coarse signal for structured head removal,
    especially at moderate retention, but it is not a universally optimal
    pruning rule. Strong single-metric baselines can outperform it at aggressive
    sparsity, and one critical Gemma-7B head produces a distinct failure mode
    that must be controlled explicitly.

\section{Theory}
\label{sec:theory}

We consider the classical multi-head attention mechanism \cite{vaswani2023attentionneed}. 
Let $X \in \mathbb{R}^{L\times d}$ denote the matrix of token representations with rows $x_i^\top$. 
Queries, keys, and values are obtained via linear projections $Q=XW^Q$, $K=XW^K$, $V=XW^V$. 
For a single head, attention computes a weighted sum of values:
\[
    \mathrm{Attn}(Q,K,V)=\mathrm{softmax}\!\left(\frac{QK^\top}{\sqrt{d_k}}\right)V,
\]
and multi-head outputs are concatenated and linearly projected (via $W^O$). 
This construction yields a row-stochastic attention matrix, placing attention weights on the probability simplex and thereby inducing a non-trivial geometry over token embeddings.

While full-token aggregation is effective, prior work shows it can collect diffuse or redundant information \cite{brunner2020identifiabilitytransformers,choromanski2022rethinkingattentionperformers,katharopoulos2020transformersrnnsfastautoregressive}. 
In our study we analyse the \emph{selection} induced by the top-$N$ attention weights and quantify separability in value–state space. Given attention weights $\{\alpha_i\}_{i=0}^L$ and corresponding value vectors $\{v_i\}_{i=0}^L$, 
let $I_N$ denote the indices of the top-$N$ tokens by attention weight. 
We define the \emph{representative vector} as
\begin{equation}
    s = \sum_{i \in I_N} \alpha_i v_i .
\end{equation}
This induces an inner geometry over tokens in which each head and layer performs information-based selection. 
Selection-based aggregation is conceptually related to sparse and structured attention mechanisms \cite{martins2016softmaxsparsemaxsparsemodel,peters2019sparsesequencetosequencemodels,niculae2019regularizedframeworksparsestructured}, as well as recent empirical studies on sink tokens and token filtering \cite{shin2025orthoranktokenselectionsink,gu2025attentionsinkemergeslanguage,guo2024activedormantattentionheadsmechanistically}. 
Unlike these approaches, our focus is the \emph{geometric separability} of the selected tokens, which we quantify using adapted classification metrics and analyse both theoretically and empirically.

\begin{figure*}[!h]
    \centering
    \includegraphics[width=0.9\linewidth]{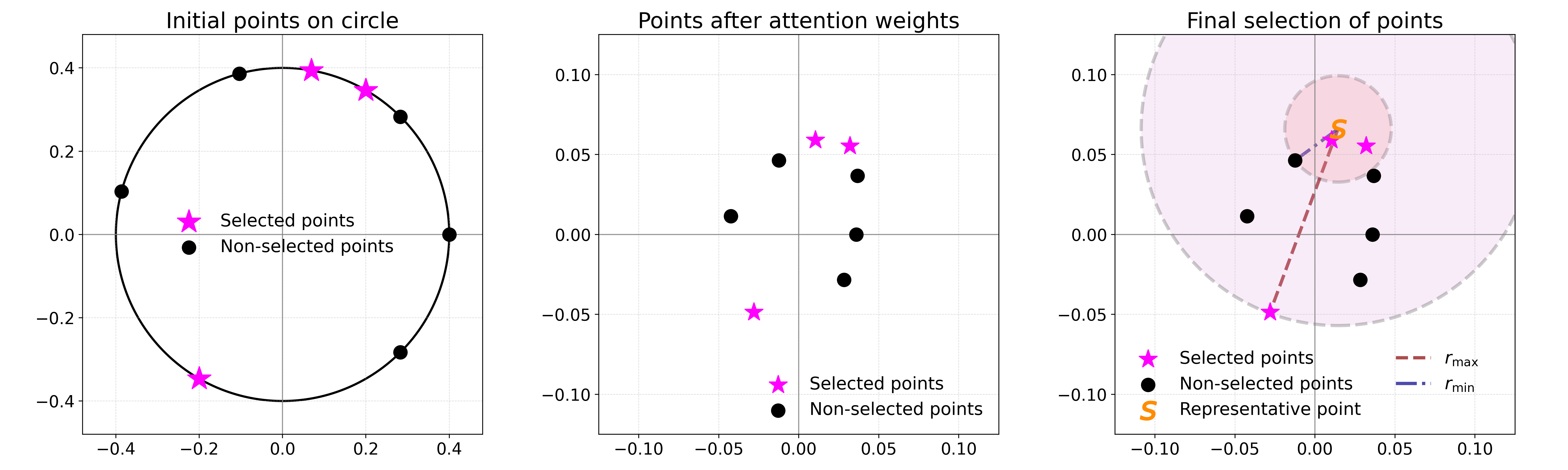}
    \caption{\textbf{Illustrative 2D example of geometric separability in value-state space.}
    \textbf{Left:} Token value-state embeddings lie on a circle.
    \textbf{Middle:} After scaling by their attention weights~$\alpha_i$, the corresponding contribution vectors $y_i=\alpha_i v_i$ move toward the origin. Selected top-$N$ contributions are shown as magenta stars, and the remaining contributions as black dots.
    \textbf{Right:} Geometric precision and recall measure how well the top-$N$ contribution vectors separate from the rest around the top-$N$ aggregate $s_N$. The radii $r_{\min}$ and $r_{\max}$ illustrate extremal neighborhoods used in the analysis.}
    \label{fig:geom_expl}
\end{figure*}

%\section{Theory}
%\label{sec:theory}

We focus on the geometry induced by multi-head attention in value-state space.
Consider attention weights $\{\alpha_i\}_{i=0}^{L-1}$ and corresponding value-state embeddings
$\{v_i\}_{i=0}^{L-1}$ for a fixed query position, layer, and head. Rather than treating
attention weights as direct indicators of semantic relevance, we study a more limited and
explicitly geometric question: whether the top-attended value contributions form a coherent
and separable set in value-state space.

Let $y_i = \alpha_i v_i$
% \begin{equation}
%     y_i = \alpha_i v_i
% \end{equation}
denote the attention-scaled value contribution of token $i$. For a fixed selection size $N$,
let $I_N \subseteq \{0,\dots,L-1\}$ be the set of indices corresponding to the top-$N$
attention weights, where $1 \leq N \leq L$. We define the top-$N$ contribution aggregate as
\begin{equation}
    s_N = \sum_{i \in I_N} y_i
        = \sum_{i \in I_N} \alpha_i v_i .
    \label{eq:topn_prototype}
\end{equation}
We use the unnormalized aggregate $s_N$ rather than a normalized centroid because $s_N$ is the
object induced by attention and preserves the attention-dependent scaling of the contribution
space.

Our goal is to quantify how well the selected top-$N$ contribution vectors separate from the
remaining contribution vectors in value-state space. The resulting metrics are diagnostic
measures of geometric separability; they should not be interpreted as ground-truth semantic,
causal, or task relevance labels.

\begin{definition}[Geometric precision and recall]
Let $B_r(s_N)$ be a ball of radius $r$ centered at the top-$N$ aggregate $s_N$.
The \textbf{geometric precision} and \textbf{geometric recall} of selection $I_N$ are
\begin{equation}
    \label{eq:precision}
    P(r,N)=
    \frac{
        \left|\{i\in I_N : y_i \in B_r(s_N)\}\right|
    }{
        \left|\{i : y_i \in B_r(s_N)\}\right|
    },
\end{equation}
and
\begin{equation}
    \label{eq:recall}
    R(r,N)=
    \frac{
        \left|\{i\in I_N : y_i \in B_r(s_N)\}\right|
    }{N}.
\end{equation}
\end{definition}

Precision measures the fraction of contribution vectors near the top-$N$ aggregate that come
from the top-$N$ selected set. Recall measures the fraction of selected top-$N$ contribution
vectors that lie near the aggregate. Importantly, ``selected'' means selected by top-$N$
attention weight, not semantically relevant.

Since the full dependence on $r$ is analytically intractable, we consider two extremal choices:
\begin{equation}
    r_{\min} = \min_{j \notin I_N} \|y_j-s_N\|_2,
    \qquad
    r_{\max} = \max_{i \in I_N} \|y_i-s_N\|_2 .
\end{equation}
By construction,
\begin{equation}
    P(r_{\min},N) \equiv 1,
    \qquad
    R(r_{\max},N) \equiv 1.
\end{equation}
Thus the nontrivial quantities are $P(r_{\max},N)$ and $R(r_{\min},N)$.

To summarize both types of geometric error with a single number, we define the extremal geometric F-score
\begin{equation}
    \label{eq:fscore}
    F_N =
    \frac{
        2P(r_{\max},N)R(r_{\min},N)
    }{
        P(r_{\max},N)+R(r_{\min},N)
    }.
\end{equation}
This score is high only when the selected neighborhood is both pure and coherent. 
Precision controls contamination by unselected tokens, while recall controls whether selected tokens remain close to their aggregate. 
Thus $F_N$ provides a compact measure of selected-unselected geometric separability.

We use a sink-aware constant-norm approximation, motivated by the empirical evidence in Appendix~\ref{ap:value_norms}-\ref{ap:value_cosines}. 
For a fixed layer $l$ and head $h$, non-sink value states are assumed to have a shared norm,
\begin{equation}
    v_i^{(l,h)} = C_{l,h}u_i^{(l,h)}, 
    \qquad \|u_i^{(l,h)}\|_2=1,
    \qquad i>0 .
\end{equation}
The first token is treated separately as an attention sink~\cite{xiao2023efficient} with smaller norm,
\begin{equation}
    v_0^{(l,h)}=\lambda_{l,h}C_{l,h}u_0^{(l,h)},
    \qquad \|u_0^{(l,h)}\|_2=1,
    \qquad 0<\lambda_{l,h}<1 .
\end{equation}

We also assume bounded directional coherence of normalized value states. 
For non-sink tokens,
\begin{equation}
    |\langle u_p,u_q\rangle|\leq \mu_N,
    \qquad p,q>0,\quad p\neq q .
\end{equation}
For the sink direction, we use a signed bound reflecting its mostly negative alignment with non-sink directions:
\begin{equation}
    -\nu_N^{+}
    \leq
    \langle u_0,u_k\rangle
    \leq
    -\nu_N^{-},
    \qquad k>0,
    \qquad
    0\leq \nu_N^{-}\leq \nu_N^{+}\leq 1 .
\end{equation}
The constants $\mu_N,\nu_N^{-},\nu_N^{+}$ are worst-case quantities; robust quantile versions are reported empirically in the appendix.

Finally, define effective weights
\begin{equation}
    \beta_i =
    \begin{cases}
        \alpha_i, & i>0,\\
        \lambda_{l,h}\alpha_0, & i=0.
    \end{cases},
% \end{equation}
% Then
% \begin{equation}
    \qquad y_i=\alpha_i v_i = C_{l,h}\beta_i u_i .
\end{equation}

% \subsection{Sink-aware deterministic bounds}

We use the empirical observations from Appendix~\ref{ap:value_norms}-\ref{ap:value_cosines} to analyze a sink-aware constant-norm regime. 
For a fixed layer $l$ and head $h$, assume that non-sink value states have constant norm,
\begin{equation}
    v_i^{(l,h)} = C_{l,h}u_i^{(l,h)}, 
    \qquad \|u_i^{(l,h)}\|_2=1,
    \qquad i>0,
\end{equation}
while the sink token has a separate smaller scale,
\begin{equation}
    v_0^{(l,h)}=\lambda_{l,h}C_{l,h}u_0^{(l,h)},
    \qquad 0<\lambda_{l,h}<1 .
\end{equation}
We also assume bounded directional coherence among normalized value directions, with a separate signed coherence condition for the sink; precise definitions are given in Appendix~\ref{ap:sink_bounds}. 
These assumptions are supported empirically by value-norm concentration and cosine statistics in Appendix~\ref{ap:value_norms}-\ref{ap:value_cosines}.

Let $S=I_N$ be the top-$N$ attention set and define
\begin{equation}
    D_i=\|y_i-s_N\|_2^2,
    \qquad
    s_N=\sum_{k\in S}y_k .
\end{equation}
An inversion is a pair $(i,j)$ with $i\in S$, $j\notin S$, and
$ D_j\leq D_i $.

The sink-aware coherence assumptions imply that only a subset of selected-unselected pairs can become inversions. 
Let $K_N^{\mathrm{sink}}$ denote the number of such potentially bad pairs. 
It is computed from attention weights, the sink scale $\lambda_{l,h}$, and the coherence parameters; its explicit form is given in Appendix~\ref{ap:sink_bounds}.

\begin{theorem}[Sink-aware deterministic precision and recall]
\label{thm:sink_aware_bound}
For a fixed layer, head, and query position, suppose the sink-aware constant-norm and coherence assumptions hold. 
Then
\begin{equation}
    P(r_{\max},N)
    \geq
    \frac{1}{1+K_N^{\mathrm{sink}}/N},
% \end{equation}
% and
% \begin{equation}
    \qquad R(r_{\min},N)
    \geq
    \max\left\{0,1-\frac{K_N^{\mathrm{sink}}}{N}\right\}.
\end{equation}
\end{theorem}

The theorem shows that precision and recall are high whenever the number of potentially bad selected-unselected pairs is small compared to $N$. 
The sink token affects the bound only through its effective weight $\lambda_{l,h}\alpha_0$ and its signed alignment with non-sink directions. 
Thus, the sink can either reduce intrusions when it is unselected or distort the selected class when it is included in $S$.

\section{Empirical study}
\label{sec:experiments}

% Our experiments test the geometric-classifier view of attention from five complementary angles. 
% First, we verify that top-$N$ attention aggregation produces a meaningful selected class: top-attention subsets should have higher geometric precision and recall than random subsets of the same size. 
% Second, we test whether the quantities appearing in our theoretical bounds, especially inversion counts and coherence-based terms, predict the observed classifier quality. Third, we isolate the attention sink and measure how its different norm and signed cosine alignment affect the geometry of selected and unselected tokens. 
% Fourth, we use the sink behavior to build a simple taxonomy of attention heads and study how different head types affect geometric classification.  
% Finally, we evaluate whether the geometry can be used practically for sparsification: if the classifier view is useful, low-budget or top-attention selections should preserve attention outputs better than random or purely attention-mass-based baselines.

We evaluate the proposed geometric diagnostics on  models from the LLaMA, Gemma, and Mistral families on two text corpora: WikiText-103 and OpenWebText~\cite{Gokaslan2019OpenWeb}. Unless otherwise stated, sequences are truncated to context length $1024$, and all metrics are computed at the final query position for every layer and attention head. 
%All experiments are run on a single NVIDIA RTX A6000 GPU with 48GB of memory.

\subsection{Top-attention sets are geometrically separable}

We first test whether top-$N$ attention aggregation defines a meaningful geometric class. For each layer, head, and query position, we take the top-$N$ attention tokens as the selected set $I_N$ and compare it with a random subset of the same size. We then compute geometric precision, recall, and F-score in contribution space. 

\begin{figure}[!h]
    \centering

    \begin{subfigure}[t]{0.3\linewidth}
        \centering
        \begin{overpic}[width=\linewidth]{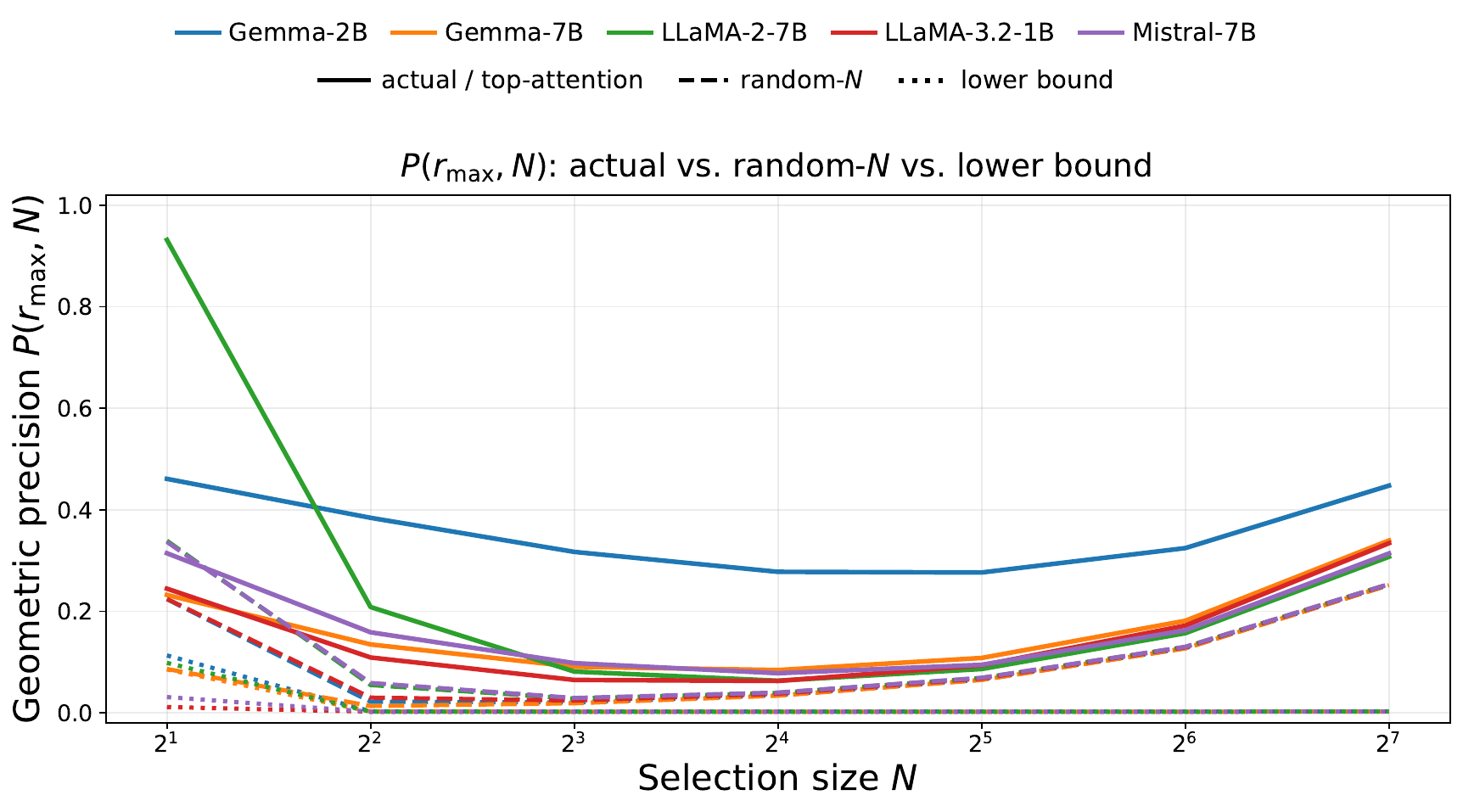}
            \put(2,45){\textbf{(a)}}
        \end{overpic}
        \phantomcaption
        \label{fig:precision_actual_random_bound}
    \end{subfigure}
    \hfill
    \begin{subfigure}[t]{0.3\linewidth}
        \centering
        \begin{overpic}[width=\linewidth]{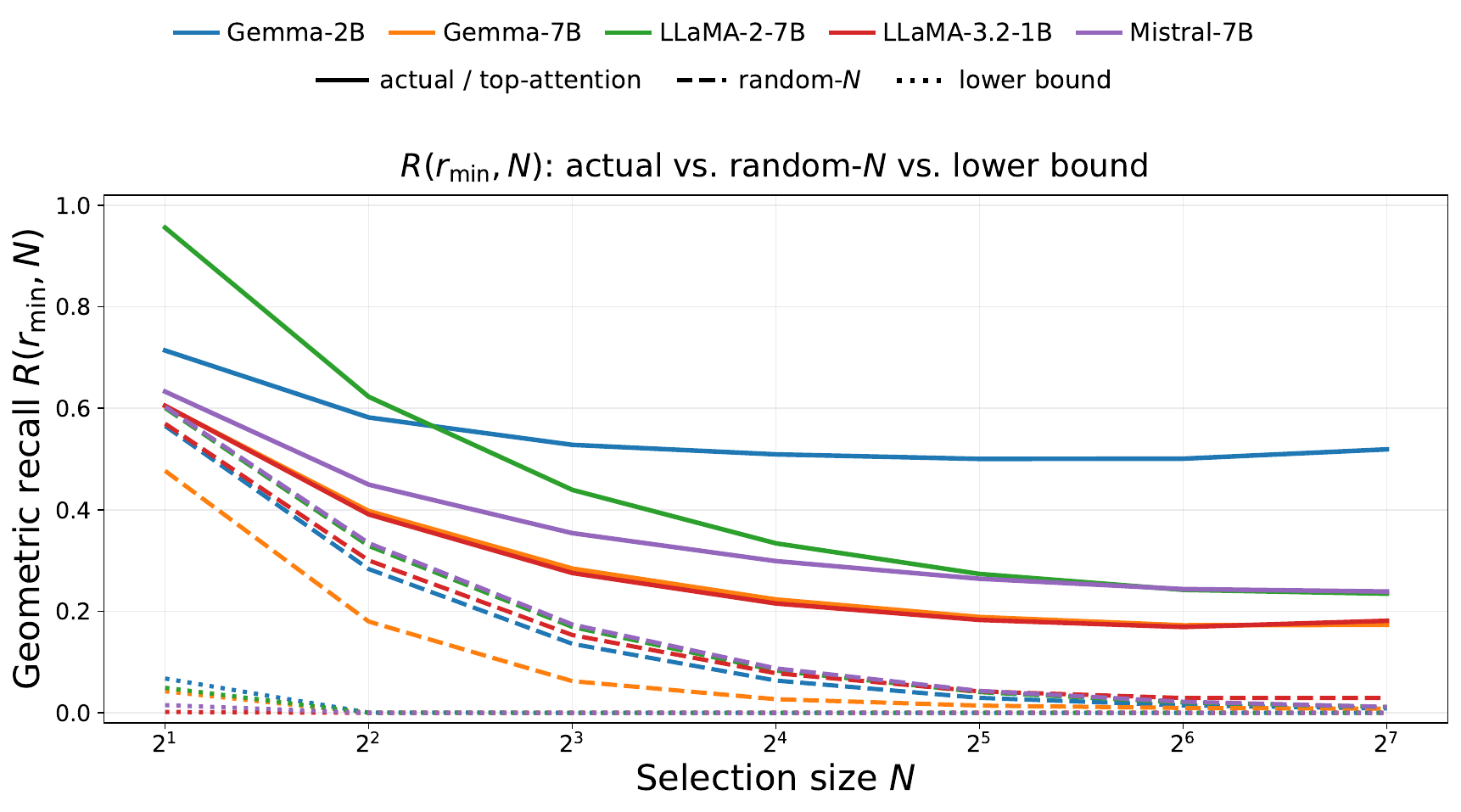}
            \put(2,45){\textbf{(b)}}
        \end{overpic}
        \phantomcaption
        \label{fig:recall_actual_random_bound}
    \end{subfigure}
    \hfill
    \begin{subfigure}[t]{0.3\linewidth}
        \centering
        \begin{overpic}[width=\linewidth]{Image/combined_recall_openwebtext_L512_actual_newrandom_oldbound.pdf}
            \put(2,45){\textbf{(b)}}
        \end{overpic}
        \phantomcaption
        \label{fig:recall_actual_random_bound_copy}
    \end{subfigure}

    \caption{
    \textbf{Geometric precision/recall, random controls, and bound quality.}
    \textbf{(a)} Geometric precision $P(r_{\max},N)$ as a function of selection size $N$.
    Solid curves denote the empirical metric for top-attention selection, dashed curves denote random-$N$ controls, and dotted curves denote the sink-aware lower bound.
    \textbf{(b)} Geometric recall $R(r_{\min},N)$ under the same comparison.
    Across models, top-attention selection consistently outperforms random selection, while the theoretical lower bounds remain valid but conservative, especially for larger $N$.
    }
    \label{fig:geometry_and_bound_quality}
\end{figure}

Figure~\ref{fig:geometry_and_bound_quality} shows that top-attention selection consistently induces stronger geometric separability than random-$N$ selection across all considered models. 
The gap is especially large for small retrieval sizes ($N<16$) where random subsets quickly lose separability, while top-attention subsets remain comparatively concentrated around their aggregate contribution vector.  

The recall curves in Fig.~\ref{fig:recall_actual_random_bound} provide the clearest evidence of this effect. A nontrivial fraction of the top-attention contribution vectors remains within the tight neighborhood of the aggregate $s_N$, while randomly selected vectors quickly fall outside the corresponding neighborhood. This indicates that attention-selected contributions are not only ranked highly by weight, but also geometrically coherent in value-state space. The precision curves in Fig.~\ref{fig:precision_actual_random_bound} show a weaker separation as top-attention selection generally remains above the random baseline, although the margin decreases for moderate and large $N$.

Overall, results suggest that attention tends to retrieve value-state contribution vectors that are more mutually aligned than vectors selected at random. As a result, the top-attended vectors contribute to the attention output in a partially coherent direction, reflected by their stronger concentration around the aggregate vector $s_N$. However, the recall curves show that only a subset of the selected vectors lies within the tightest neighborhood of $s_N$. This indicates that the selected vectors do not form a single compact cluster. Instead, the top-attended set appears to contain both a geometrically central subset, which is closely aligned with the aggregate, and a more peripheral subset, whose vectors are farther from $s_N$ and may contribute complementary or context-specific directions. At the same time, the relatively low precision values indicate that the neighborhood around $s_N$ is not occupied exclusively by selected vectors. Many nearby non-selected vectors are geometrically similar to the selected ones but do not receive top attention. Thus, attention induces a statistically enriched and partially coherent region in value-state space, rather than a cleanly separated cluster of selected vectors.

We next evaluate how informative the sink-aware deterministic bounds are in practice. 
For each model, we compare the actual geometric precision and recall with the lower bounds from Theorem~\ref{thm:sink_aware_bound}. 
The bounds use only attention weights, the sink scale, and coherence quantities, and therefore are expected to be conservative.

Figures~\ref{fig:precision_actual_random_bound} and~\ref{fig:recall_actual_random_bound} show that the sink-aware bounds correctly behave as lower bounds, but are loose for most non-trivial \(N\). 
This is because \(K_N^{\mathrm{sink}}\) counts all pairs that \emph{could} become inversions under worst-case coherence, even though only a subset of them actually invert. 
Thus the theorem should be interpreted as a conservative certificate of separability, not as a tight estimator of precision or recall.

%The empirical results support the validity of the deterministic bound, while also showing that worst-case coherence is too conservative for quantitative prediction. 
%This motivates reporting robust empirical coherence statistics in the appendix and using the theorem mainly as an explanatory tool for how sink scale and directional alignment affect geometric separability.

\subsection{Attention sinks induce a distinctive geometry in value-contribution space}
\label{sec:sink_effect}

We next isolate the role of the first token. 
The sink-aware theory assumes that the first token has a distinct value-state norm and a different directional relationship to non-sink tokens. 
We verify both properties empirically and then test whether the sink changes the geometry of the selected top-\(N\) class.

\begin{figure}[h]
    \centering

    % -------- left column --------
    \begin{minipage}[c]{0.35\linewidth}
        \centering

        \begin{subfigure}[t]{\linewidth}
            \centering
            \begin{overpic}[width=0.95\linewidth]{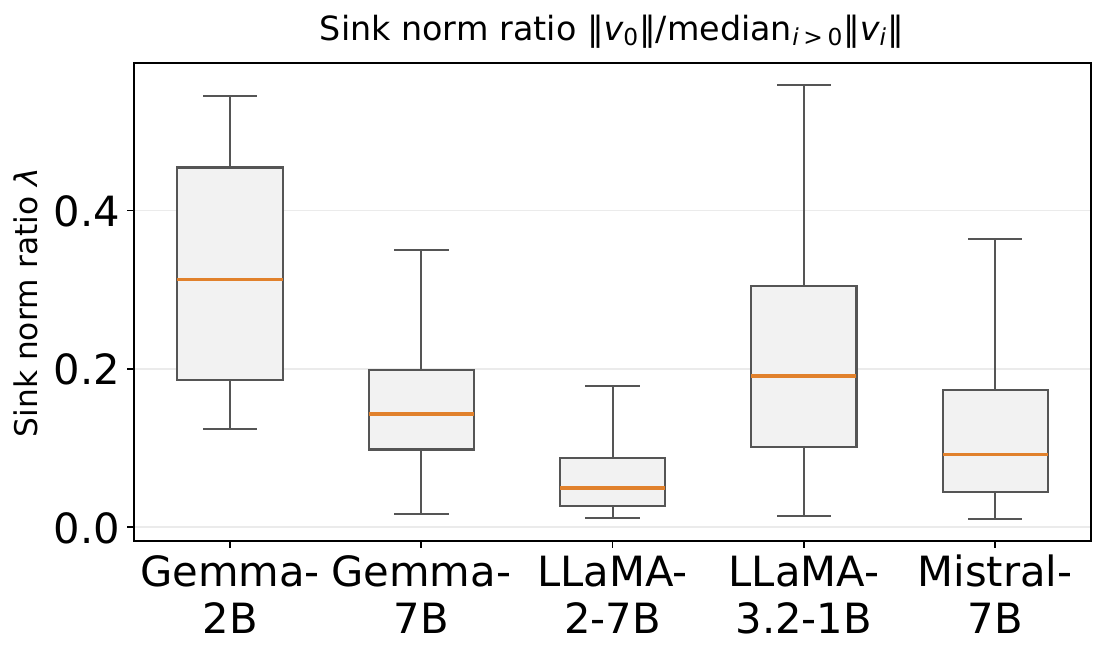}
                \put(2,55){\textbf{(a)}}
            \end{overpic}
            \phantomcaption
            \label{fig:sink_norm_ratio}
        \end{subfigure}

        \vspace{0.9em}

        \begin{subfigure}[t]{\linewidth}
            \centering
            \begin{overpic}[width=0.95\linewidth]{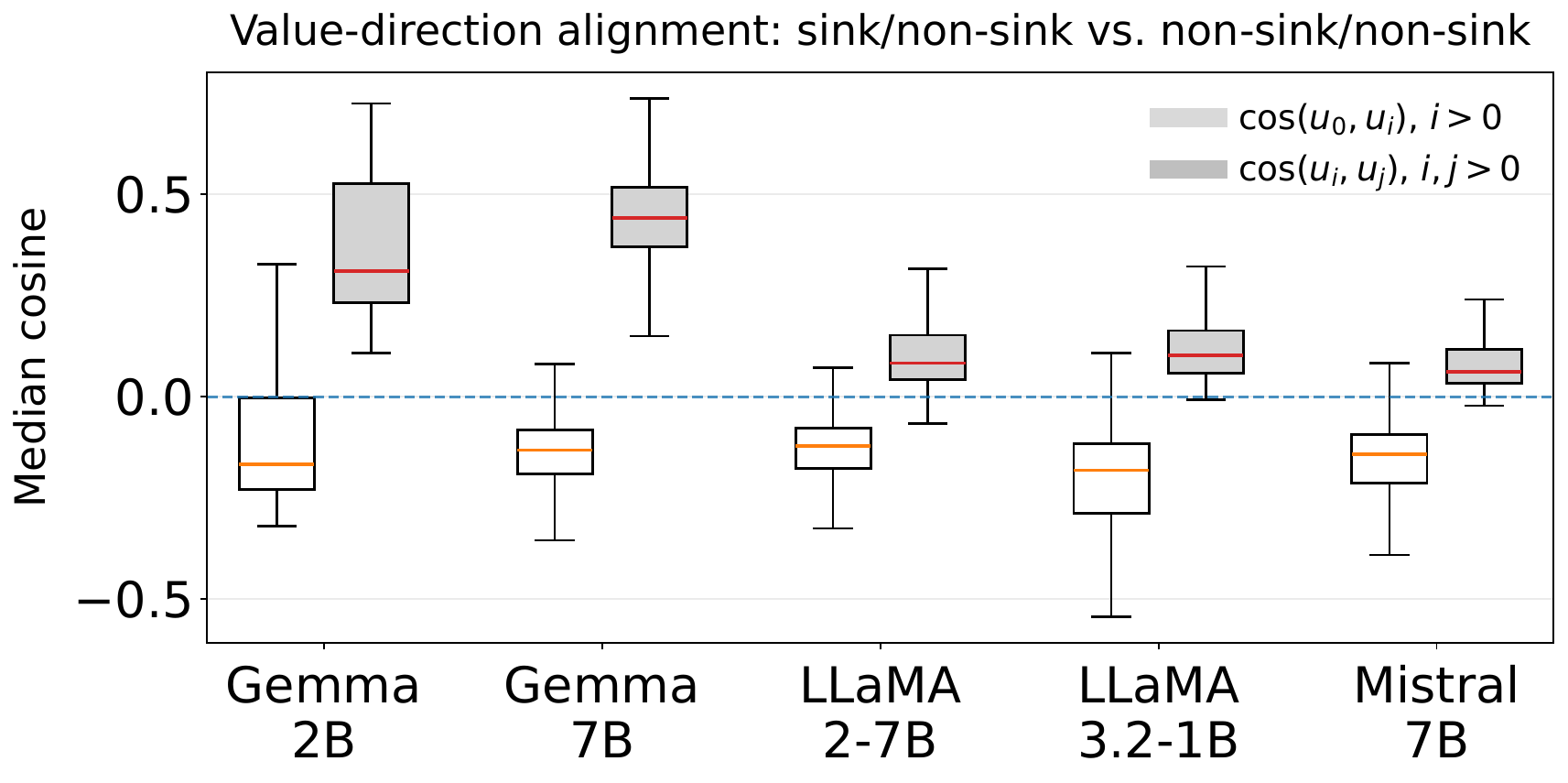}
                \put(2,55){\textbf{(b)}}
            \end{overpic}
            \phantomcaption
            \label{fig:sink_alignment}
        \end{subfigure}

    \end{minipage}
    \hfill
    % -------- right column --------
    \begin{minipage}[c]{0.62\linewidth}
        \centering

        \begin{subfigure}[t]{\linewidth}
            \centering
            \begin{overpic}[width=\linewidth]{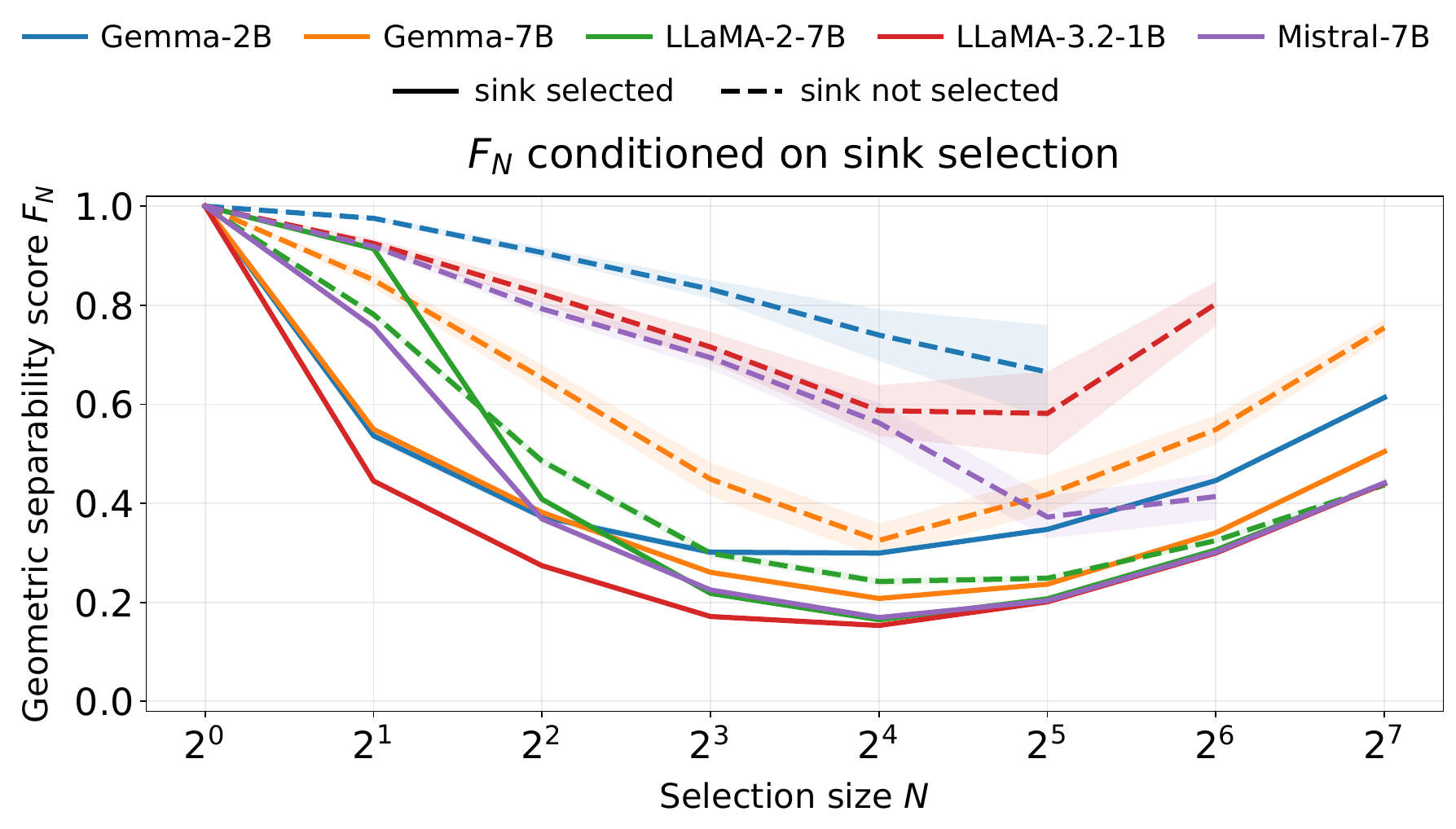}
                \put(2,50){\textbf{(c)}}
            \end{overpic}
            \phantomcaption
            \label{fig:sink_exclusion_effect}
        \end{subfigure}

    \end{minipage}

    \caption{
    \textbf{Sink tokens have outlier value-state geometry and reduce top-$N$ separability.}
\textbf{(a)} Sink values have substantially smaller norms than typical non-sink values.
    \textbf{(b)} Despite their small norm, sink values are directionally atypical because sink--non-sink value cosines are slightly negative, whereas non-sink value pairs are approximately orthogonal or positively aligned.
    \textbf{(c)} Top-$N$ attention-selected contributions are less geometrically separable when the sink is included. The drop in $F_N$ indicates that attention sinks disrupt the coherence of the top-attended set, especially in sparse top-$N$ regimes. Solid curves with sink, dashed curves without sink.
    }
    \label{fig:sink_geometry_and_effect}
\end{figure}

Prior work on attention sinks has primarily characterized them as key-side attractors that receive large attention mass even when they are not semantically salient. This view predicts that sink values should not behave like ordinary content-bearing value vectors, since a large semantic value attached to a broadly attended sink would be injected into many query positions. 
Figure~\ref{fig:sink_geometry_and_effect}(a) reports the sink-to-median norm ratio $\lambda$ across
five open-weight models spanning three families (Gemma, LLaMA, Mistral) and a $7\times$
parameter range. Across all models, the median $\lambda$ falls in the $0.05$--$0.30$
range, confirming that the value vector at the sink position is $3$--$10\times$ smaller
in norm than a typical value vector in the same head. This extends prior single-model
observations of value-state drain~\citep{bondarenko2023quantizable, guo2024active,
sun2024massive} to a broader architectural and scale range, and supports the view that
suppressed sink-value norm is a stable, training-induced property of softmax-attention
transformers rather than an artifact of any specific tokenizer, training corpus, or
model size.

%Figure~\ref{fig:sink_geometry_and_effect} shows that the sink token is geometrically different from ordinary tokens. In the models included in our study norms of value tokens are almost constant (see App.~\ref{ap:value_norms}, fig.~\ref{fig:ap_value_norms}).

% Sink value-state norm is substantially smaller than the typical non-sink norm (fig.~\ref{fig:sink_geometry_and_effect}a), supporting the parametrization
% \[
%     \|v_0^{(l,h)}\|_2 = \lambda_{l,h} C_{l,h},
%     \qquad \lambda_{l,h}<1.
% \]

However, the sink value is not simply an isotropic near-zero vector. Figure~\ref{fig:sink_geometry_and_effect}(b) characterizes the \emph{directional} structure of
value vectors, which has been comparatively under-examined in prior work. Pairwise
cosines between non-sink value directions are near-orthogonal in LLaMA and Mistral
(median $\approx 0.1$) and weakly co-aligned in Gemma (median $\approx 0.4$),
reflecting a head-specific modal direction along which content tokens contribute.
Cosines between the sink direction and non-sink directions, in contrast, are
systematically negative (median $\approx -0.1$ to $-0.2$) across all five models. The
sink is therefore not merely a near-origin point in value space but its residual
orientation is consistently \emph{opposed} to the direction of content values. This is consistent with, and provides direct geometric
evidence for, the ``scaling-factor-within-softmax'' or implicit-rescaling
interpretation of sinks~\citep{cancedda2024spectral, an2025unified}, in which the sink
contribution implements a learned subtraction along the head's modal content axis
rather than a pure no-op.

% This opposite alignment of the sink to non-sink values motivates the signed sink-coherence assumption used in Theorem~\ref{thm:sink_aware_bound}.

We next ask whether the top-attended tokens form a coherent set in attention-scaled value-contribution space. Figure~\ref{fig:sink_geometry_and_effect}(c) translates the input-side geometry of (a) and (b)
into an output-side behavioral consequence. Conditioning on whether the sink token is
among the top-$N$ attention-selected positions, we observe a large and consistent gap
in $F_N$ across all five models: including the sink reduces geometric separability by a
factor of roughly two (e.g., LLaMA-3.2-1B drops from $F_N \approx 0.6$ to $\approx
0.15$ at $N = 16$). The gap is largest in the sparse-but-nontrivial regime $N \in
[2^3, 2^5]$, which is precisely the operating range of top-$K$ attention pruning,
sparse-attention kernels, KV-cache eviction policies, and attention-rollout
interpretability methods. The dashed (sink-excluded) curves themselves remain below
$1$, indicating that top-$N$ attention weight is an imperfect proxy for value-space
coherence even in the absence of a sink, but the sink-included curves show that the
dominant disruption comes from a single token.

Together, these
results support a unified picture of the sink as a \emph{directionally anomalous,
magnitude-suppressed} sub-computation that shares the head's softmax with content
tokens but does not belong to their geometric population. Methods that treat top-$N$
attention weights as a homogeneous relevance signal therefore make a category error in
exactly the regime where they are most often deployed.

\subsection{Geometric head taxonomy}
\label{sec:head_taxonomy}

We define the regimes with the same source-scaled contribution score used in
our top-\(N\) analysis. For sequence \(m\), layer \(\ell\), head \(h\), and
query position \(i>0\), let
\begin{equation}
    j^{\star}_{m,\ell,h,i}
    =
    \arg\max_{0\leq j\leq i}
    A^{(m,\ell,h)}_{ij}\,\lVert v^{(m,\ell,h)}_j\rVert_2.
    \label{eq:source_winner}
\end{equation}
The winning source is labelled \emph{Diagonal} when \(j^\star=i\),
\emph{Sink} when \(j^\star=0\), and \emph{Other} otherwise. We first take a
plurality vote over query positions within each sequence and then a plurality
vote over sequences. Exact ties are mapped to \emph{Other}, the conservative
choice indicating that no single privileged source is established. Position
\(i=0\) is excluded because sink and diagonal coincide there. The final,
parameter-free semantic mapping is
\begin{equation}
    \text{Diagonal}\mapsto\textsc{Retriever},\qquad
    \text{Sink}\mapsto\textsc{Mixer},\qquad
    \text{Other}\mapsto\textsc{Reset}.
    \label{eq:article_taxonomy_mapping}
\end{equation}
Algorithm~\ref{alg:head_classification} gives the full procedure.

\textbf{Retriever.}
The current token is the most frequent contribution winner. These heads retain
strong alignment with \(v_{L-1}\), and removing the current-token contribution
changes the selected aggregate more than removing the sink.

\textbf{Mixer.}
The sink is the most frequent contribution winner. Their aggregate remains
sink-aligned at small and intermediate \(N\), while alignment with the current
token increases as more content contributions enter the selection.

\textbf{Reset.}
Neither the sink nor the current token wins most often. The winning source is
distributed over non-privileged context positions, and the aggregate is
therefore comparatively insensitive to either single leave-one-out removal.

\begin{figure}[t]
    \centering
    \includegraphics[width=\linewidth]{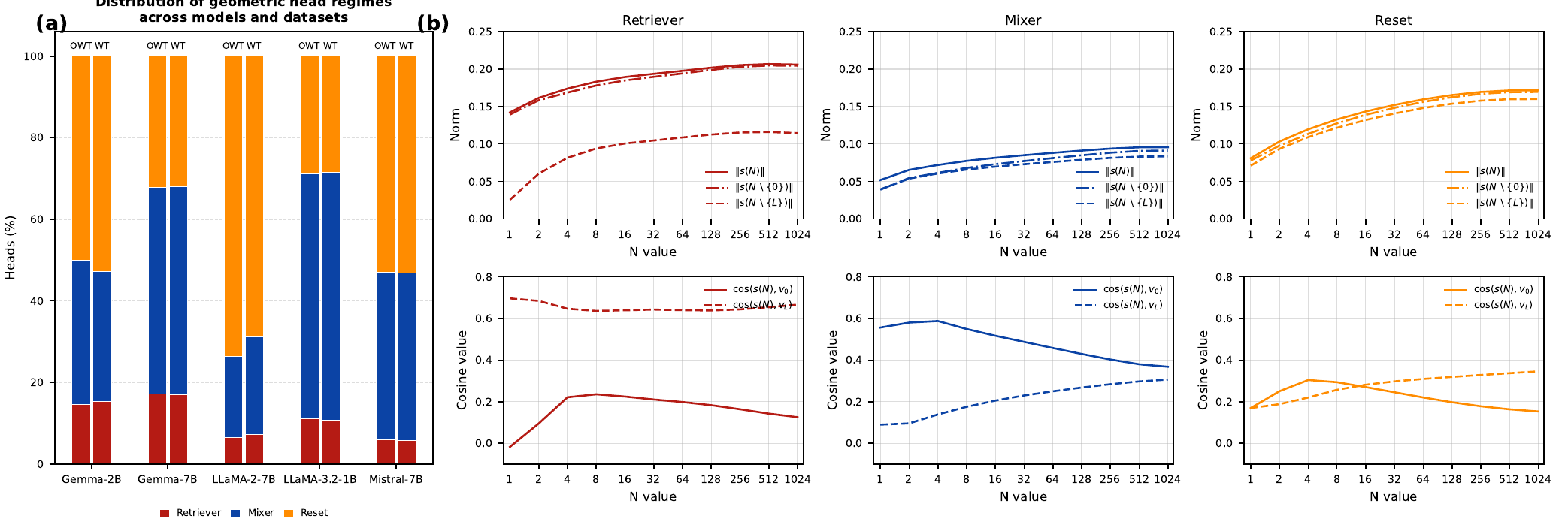}
    \caption{\textbf{A parameter-free geometric taxonomy of attention heads.}
    \textbf{(a)} Per-model distributions on OpenWebText (OWT) and WikiText
    (WT), using equalized single-document contexts with an initial BOS token.
    Retriever is the minority regime for every model (\(5.8\)--\(17.2\%\)).
    Mixer is largest for Gemma-7B and LLaMA-3.2-1B, whereas Reset is largest
    for LLaMA-2-7B, Mistral-7B, and Gemma-2B; corresponding OWT--WT fractions
    differ by less than five percentage points.
    \textbf{(b)} Mechanistic signatures for LLaMA-2-7B, averaged within each
    regime. \emph{Top:} normalized aggregate norm \(\lVert s(N)\rVert\)
    (solid) and leave-one-out variants removing the sink (dash-dot) or current
    token (dashed). \emph{Bottom:} alignment with \(v_0\) (solid) and
    \(v_{L-1}\) (dashed). Retriever heads remain current-token aligned; Mixer
    heads retain stronger sink alignment while current-token alignment grows
    with \(N\); Reset heads show modest, converging alignments and nearly
    overlapping leave-one-out norms.}
    \label{fig:head_taxonomy}
\end{figure}

Figure~\ref{fig:head_taxonomy}(a) shows that the new rule no longer imposes a
universal majority class. Retriever heads are consistently rare, ranging from
\(5.8\%\) to \(17.2\%\) across model--dataset pairs. Mixer is the largest
regime for Gemma-7B (\(50.7\)--\(51.1\%\)) and LLaMA-3.2-1B
(\(60.0\)--\(60.7\%\)). Reset is the largest regime for LLaMA-2-7B
(\(68.8\)--\(73.5\%\)), Mistral-7B (\(52.9\)--\(53.1\%\)), and Gemma-2B
(\(50.0\)--\(52.8\%\)). The largest OWT--WikiText change in any cell is below
five percentage points. Thus the assignment is reproducible across corpora,
but the balance between sink-mediated and diffuse computation is a property of
the architecture rather than a universal constant.

Figure~\ref{fig:head_taxonomy}(b) shows that the three labels also separate
mechanistic trajectories. Retriever heads maintain substantially stronger
alignment with the current-token value than with the sink value, and current-
token removal is the more consequential leave-one-out operation. Mixer heads
remain predominantly aligned with \(v_0\), while their alignment with
\(v_{L-1}\) grows steadily with selection size; this is consistent with a
sink-mediated gate that incorporates increasing amounts of content. Reset
heads have no persistent privileged direction: the two cosine trajectories are
modest and approach one another, while both leave-one-out norm curves closely
track the full aggregate. These differences arise from a parameter-free
classification based only on contribution winners, rather than from fitted
thresholds chosen to produce the observed shapes.

The taxonomy therefore refines a binary active--dormant view into a current-
token retrieval mode, a sink-mediated aggregation mode, and a diffuse reset
mode. We provide the implementation and full cross-model counts in
Appendix~\ref{ap:taxonomy}.

\subsection{Head sparsification as functional validation}
\label{sec:head_sparsification}

We next test whether the parameter-free head regimes provide a useful signal
for structured sparsification. On OpenWebText at context length \(L=1024\), we
retain the same number of heads per layer and measure the increase in next-token
negative log-likelihood, \(\Delta\mathrm{NLL}\), under a prefill plus one-token
decode protocol. The retained fraction is a structural sparsity proxy; our
current implementation does not claim wall-clock acceleration because it does
not use a dedicated sparse kernel.

For the type-guided method, regime priorities are estimated on a disjoint
importance split from exact single-head ablations. Regimes are ordered by mean
ablation impact, and heads within a regime are ordered by their individual
ablation impact. The taxonomy itself remains parameter-free: only the
functional ranking used for the downstream pruning experiment uses ablation
measurements. We compare with weight magnitude, low- and high-entropy rules,
sink- and current-token attention, and random per-layer masks.

\begin{figure}[h]
    \centering
    \includegraphics[width=\linewidth]{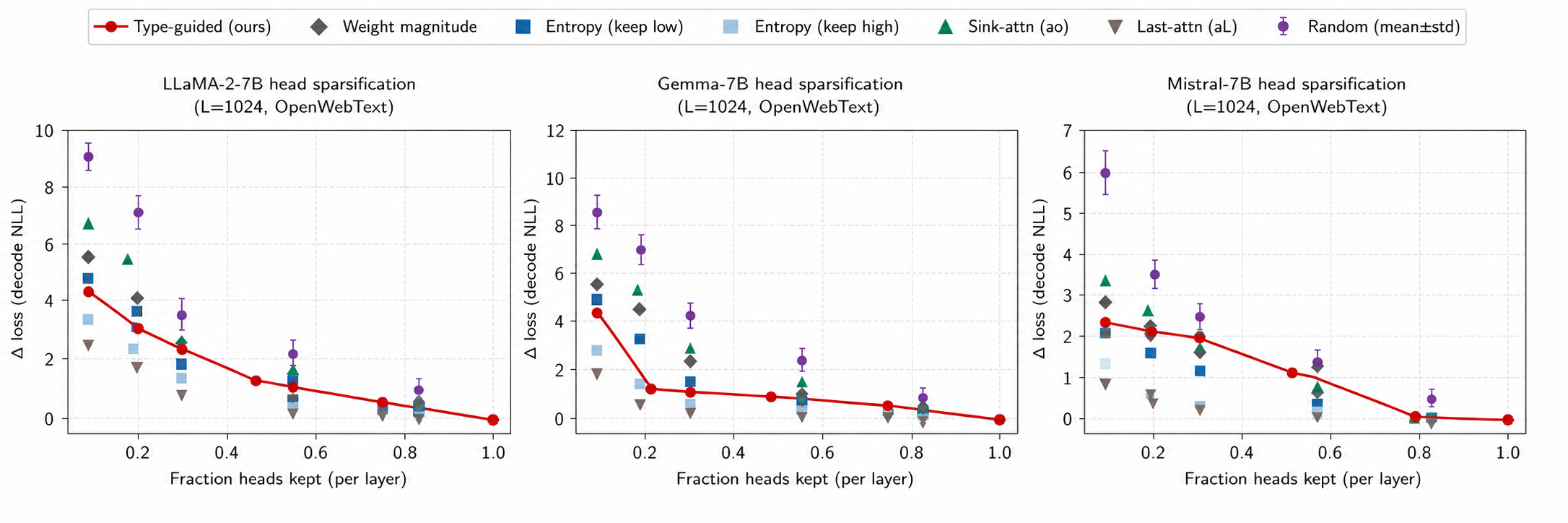}
    \caption{\textbf{Head sparsification.}
    Increase in decode NLL (\(\Delta\mathrm{NLL}\)) on OpenWebText at
    \(L=1024\) when retaining a fraction of heads per layer.
    \textbf{Type-guided (ours)} orders the parameter-free Retriever/Mixer/Reset
    groups by ablation impact and uses individual impact within each group.
    Baselines include weight magnitude, entropy (keep low/high), sink/current-
    token attention (\(a_0/a_L\)), and random removal (mean\(\pm\)std).
    For Gemma-7B, the critical head at layer 0, head 12 is held fixed in every
    mask and excluded from the removable ranking; the corresponding unfiltered
    anomaly is documented in Appendix~\ref{ap:gemma_critical_head}.}
    \label{fig:ablation}
\end{figure}

The revised results do not support a claim that type-guided sparsification is
uniformly best. For LLaMA-2-7B it is strongest at high retention, while weight,
entropy, or current-token heuristics can be better under aggressive pruning.
For Mistral-7B several single-metric and random baselines are competitive or
better at intermediate retention. The anomaly-controlled Gemma-7B result
returns to the same numerical scale as the other models and shows that the
regime ordering carries useful information, but it remains sensitive to a
small number of critical heads. We therefore interpret the taxonomy as a
coarse, architecture-dependent importance signal rather than a complete
compression rule. Robust deployment should combine regime information with
head-level sensitivity and explicit protection of critical heads.

\section{Conclusion}

We presented a geometric framework for analysing multi-head attention through
top-\(N\) selection in value-state space. Geometric precision, recall, and the
extremal score \(F_N\) expose selected--unselected inversions, while the
sink-aware theory provides conservative deterministic certificates under
empirically supported norm and coherence assumptions. Across LLaMA, Gemma, and
Mistral models, top-attention sets are more geometrically separable than random
sets and the sink token has a systematic effect on the selected aggregate.

The updated taxonomy is parameter-free. It labels each head from the plurality
of contribution-maximizing source positions and requires no thresholds,
margins, or fitted coefficients. Retriever heads are consistently a minority,
whereas the balance between sink-mediated Mixer and diffuse Reset heads varies
substantially by architecture but only modestly by corpus. Their norm and
alignment trajectories provide distinct mechanistic signatures.

Sparsification gives a more qualified functional picture than the original
analysis: regime labels are informative, particularly at moderate retention,
but they do not dominate every baseline at every sparsity level. The Gemma-7B
critical-head anomaly further shows that type-level averages can conceal rare,
individually essential heads. Geometry-aware compression should therefore use
the taxonomy as one diagnostic signal together with head-level sensitivity,
rather than as a stand-alone universal pruning rule.

\clearpage
% The bibliography is included explicitly for robust arXiv compilation.
% This avoids a BibTeX dependency in AutoTeX while preserving natbib citations.

%%%%%%%%%%%%%%%%%%%%%%%%%%%%%%%%%%%%%%%%%%%%%%%%%%%%%%%%%%%%%%%%%%%%%%%%%%%%%%%
%%%%%%%%%%%%%%%%%%%%%%%%%%%%%%%%%%%%%%%%%%%%%%%%%%%%%%%%%%%%%%%%%%%%%%%%%%%%%%%
% APPENDIX
%%%%%%%%%%%%%%%%%%%%%%%%%%%%%%%%%%%%%%%%%%%%%%%%%%%%%%%%%%%%%%%%%%%%%%%%%%%%%%%
%%%%%%%%%%%%%%%%%%%%%%%%%%%%%%%%%%%%%%%%%%%%%%%%%%%%%%%%%%%%%%%%%%%%%%%%%%%%%%%
\newpage

\appendix
\section{Appendix}

\subsection{Proof of the Sink-Aware Deterministic Bound}
\label{ap:sink_bounds}

We prove Theorem~\ref{thm:sink_aware_bound}. 
Fix a layer, head, and query position. 
For readability, we omit the superscript \((l,h)\). 
Under the sink-aware constant-norm assumption,
\begin{equation}
    y_i = \alpha_i v_i = C\beta_i u_i,
\end{equation}
where
\begin{equation}
    \beta_i =
    \begin{cases}
        \alpha_i, & i>0,\\
        \lambda \alpha_0, & i=0.
    \end{cases}
\end{equation}
Let \(S=I_N\), \(S_+=S\setminus\{0\}\), and
\begin{equation}
    A_+ = \sum_{k\in S_+}\beta_k,
    \qquad
    \sigma=\mathbf{1}\{0\in S\}.
\end{equation}

For selected \(i\in S\) and unselected \(j\notin S\), define
\begin{equation}
    D_i=\|y_i-s_N\|_2^2,
    \qquad
    X_{ij}=D_j-D_i.
\end{equation}
A pair \((i,j)\) is inverted if \(X_{ij}\leq 0\).

Expanding the squared distances gives
\begin{equation}
\begin{split}
    X_{ij}
    &=
    \|y_j-s_N\|_2^2-\|y_i-s_N\|_2^2 \\
    &=
    \|y_j\|_2^2-\|y_i\|_2^2
    -2\langle y_j-y_i,s_N\rangle .
\end{split}
\end{equation}
Substituting \(y_i=C\beta_i u_i\) and \(s_N=C\sum_{k\in S}\beta_k u_k\), we obtain
\begin{equation}
\frac{X_{ij}}{C^2}
=
\beta_i^2+\beta_j^2
+
2\beta_i
\sum_{k\in S\setminus\{i\}}
\beta_k\langle u_i,u_k\rangle
-
2\beta_j
\sum_{k\in S}
\beta_k\langle u_j,u_k\rangle .
\label{eq:ap_sink_gap_identity}
\end{equation}

We now use the coherence assumptions. 
For non-sink directions,
\begin{equation}
    |\langle u_p,u_q\rangle|\leq \mu_N,
    \qquad p,q>0,\quad p\neq q.
\end{equation}
For the sink direction,
\begin{equation}
    -\nu_N^+
    \leq
    \langle u_0,u_k\rangle
    \leq
    -\nu_N^-,
    \qquad k>0.
\end{equation}

First consider non-sink selected and non-sink unselected pairs:
\(i\in S_+\), \(j\notin S\), \(j>0\). 
From Eq.~\ref{eq:ap_sink_gap_identity},
\begin{equation}
\begin{split}
\frac{X_{ij}}{C^2}
\geq
&
\beta_i^2+\beta_j^2
-
2\mu_N
\left[
    \beta_i(A_+-\beta_i)
    +
    \beta_jA_+
\right] \\
&+
2\sigma\beta_0
\left(
    \nu_N^-\beta_j
    -
    \nu_N^+\beta_i
\right).
\end{split}
\end{equation}
Define the right-hand side as \(L_{ij}^{++}\). 
Thus,
\begin{equation}
    \frac{X_{ij}}{C^2}\geq L_{ij}^{++}.
\end{equation}

If the sink is unselected, then for \(i\in S_+\) and \(j=0\),
\begin{equation}
    \frac{X_{i0}}{C^2}
    \geq
    \beta_i^2+\beta_0^2
    -2\mu_N\beta_i(A_+-\beta_i)
    +2\beta_0\nu_N^-A_+ .
\end{equation}
Define this lower bound as \(L_{i0}^{+0}\).

If the sink is selected, then for \(i=0\) and \(j\notin S\), \(j>0\),
\begin{equation}
    \frac{X_{0j}}{C^2}
    \geq
    \beta_0^2+\beta_j^2
    -2\beta_0\nu_N^+A_+
    +2\beta_j\beta_0\nu_N^-
    -2\mu_N\beta_jA_+ .
\end{equation}
Define this lower bound as \(L_{0j}^{0+}\).

Now define \(K_N^{++}\) as the number of non-sink/non-sink pairs with
\begin{equation}
    L_{ij}^{++}\leq 0.
\end{equation}
If the sink is unselected, define \(K_N^{+0}\) as the number of selected non-sink tokens with
\begin{equation}
    L_{i0}^{+0}\leq 0.
\end{equation}
If the sink is selected, define \(K_N^{0+}\) as the number of unselected non-sink tokens with
\begin{equation}
    L_{0j}^{0+}\leq 0.
\end{equation}
Finally,
\begin{equation}
    K_N^{\mathrm{sink}}
    =
    K_N^{++}
    +
    \mathbf{1}\{0\notin S\}K_N^{+0}
    +
    \mathbf{1}\{0\in S\}K_N^{0+}.
\end{equation}

If a pair has a positive lower bound, then \(X_{ij}>0\), so the pair cannot be inverted. 
Therefore all inversions are contained in the set of pairs counted by \(K_N^{\mathrm{sink}}\). 
Let
\begin{equation}
    A_N=
    \sum_{i\in S}
    \sum_{j\notin S}
    \mathbf{1}\{D_j\leq D_i\}
\end{equation}
be the total number of selected-unselected inversions. 
Then
\begin{equation}
    A_N\leq K_N^{\mathrm{sink}}.
\label{eq:ap_A_leq_K_sink}
\end{equation}

It remains to relate inversion count to precision and recall. 
At radius \(r_{\max}\), all \(N\) selected tokens are inside the ball. 
Each unselected token inside the same ball induces at least one inversion, hence there are at most \(A_N\) such false positives. 
Therefore
\begin{equation}
    P(r_{\max},N)
    \geq
    \frac{N}{N+A_N}.
\end{equation}
Using Eq.~\eqref{eq:ap_A_leq_K_sink},
\begin{equation}
    P(r_{\max},N)
    \geq
    \frac{N}{N+K_N^{\mathrm{sink}}}
    =
    \frac{1}{1+K_N^{\mathrm{sink}}/N}.
\end{equation}

Similarly, at radius \(r_{\min}\), every selected token outside the ball induces at least one inversion with the nearest unselected token. 
Thus at most \(A_N\) selected tokens are missed, and
\begin{equation}
    R(r_{\min},N)
    \geq
    1-\frac{A_N}{N}.
\end{equation}
Using Eq.~\eqref{eq:ap_A_leq_K_sink} and clipping at zero gives
\begin{equation}
    R(r_{\min},N)
    \geq
    \max\left\{
        0,
        1-\frac{K_N^{\mathrm{sink}}}{N}
    \right\}.
\end{equation}
This proves the theorem.

\subsection{Models}

This appendix provides the model and dataset details used throughout the experiments. The main text reports only the high-level setup to save space. For the taxonomy, mechanistic-signature, and sparsification experiments, we use equalized single-document preprocessing: every context is constructed as an initial BOS token followed by one contiguous window from one document. Documents are not concatenated and no EOS token is inserted between dataset examples. WikiText rows are first reconstructed into articles before window selection. Other experiments use the preprocessing stated in their corresponding subsections.

\begin{table}[!h]
\centering
\caption{
\textbf{Models used in the experiments.}
We evaluate models from the LLaMA, Gemma, and Mistral families, covering both
small and 7B-scale checkpoints.
}
\label{tab:models}
\resizebox{\linewidth}{!}{
\begin{tabular}{llll}
\toprule
Model checkpoint & Family & Parameters & Reference \\
\midrule
\texttt{meta-llama/Llama-2-7b-hf} & LLaMA 2 & 7B & LLaMA 2~\cite{touvron2023llama2} \\
\texttt{meta-llama/Llama-3.2-1B} & LLaMA 3.2 & 1B & LLaMA 3.2 ~\cite{meta2024llama32} \\
\texttt{google/gemma-2b} & Gemma & 2B & Gemma~\cite{team2024gemma} \\
\texttt{google/gemma-7b} & Gemma & 7B & Gemma~\cite{team2024gemma} \\
\texttt{mistralai/Mistral-7B-v0.1} & Mistral & 7B & Mistral 7B~\cite{jiang2023mistral} \\
\bottomrule
\end{tabular}
}
\end{table}

\newpage
\subsection{Empirical value-norm structure}
\label{ap:value_norms}

This section supports the constant-norm approximation used in Section~\ref{sec:theory}. 
For each model, layer, and head, we measure the norms of value states
\[
    \|v_i^{(l,h)}\|_2
\]
across token positions. 
We separate the first token from the remaining positions because the first token often acts as an attention sink.

For non-sink tokens, we compute the coefficient of variation
\[
    \mathrm{CV}_{l,h}
    =
    \frac{
        \mathrm{std}_{i>0}\|v_i^{(l,h)}\|_2
    }{
        \mathrm{mean}_{i>0}\|v_i^{(l,h)}\|_2
    } .
\]
Small values of \(\mathrm{CV}_{l,h}\) indicate that non-sink value states approximately lie on a sphere of radius \(C_{l,h}\), motivating the approximation
\[
    \|v_i^{(l,h)}\|_2 \approx C_{l,h},
    \qquad i>0.
\]

\begin{figure}[!h]
    \centering
        \centering
        \includegraphics[width=\linewidth]{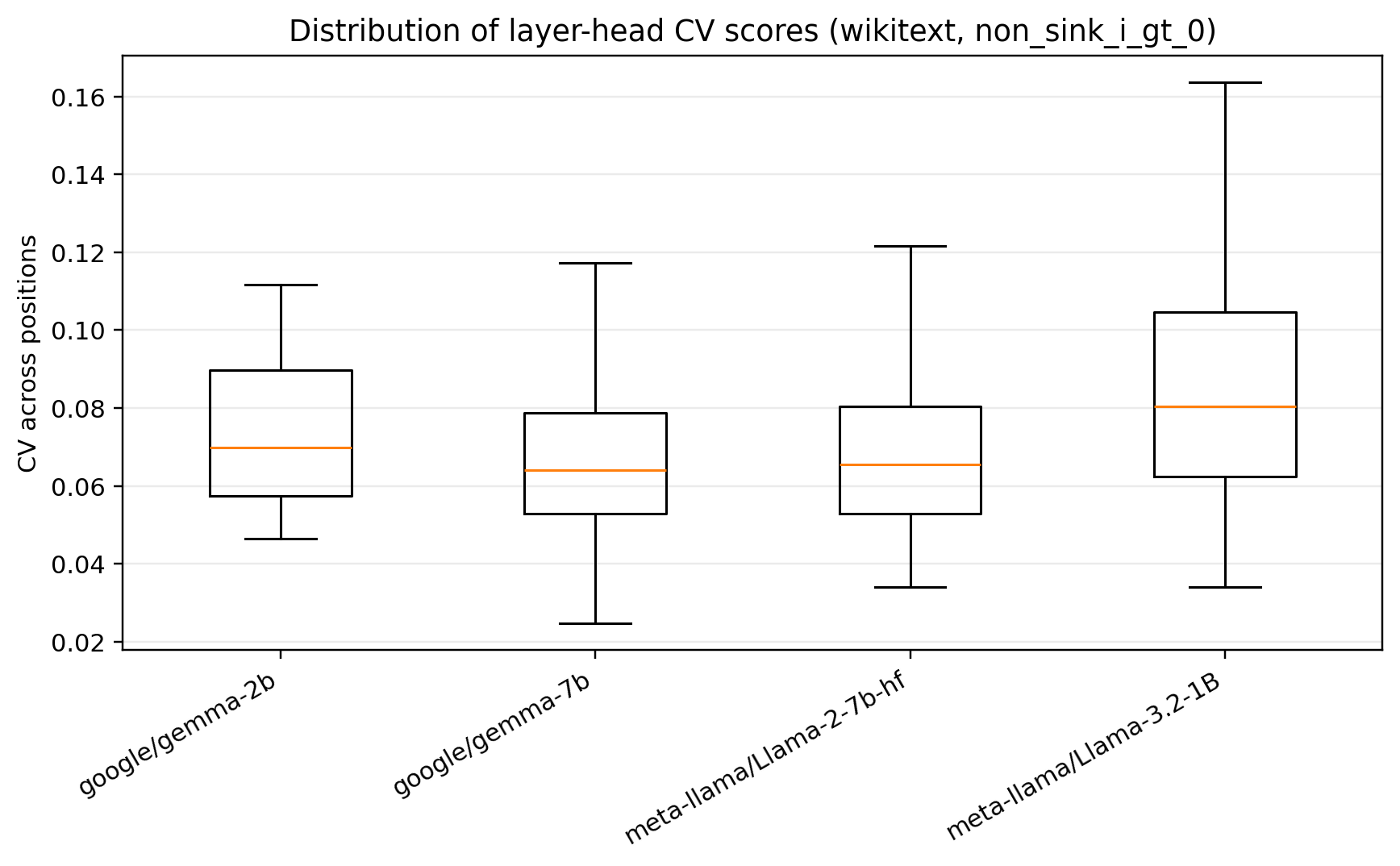}

    \caption{
    \textbf{Empirical value-norm structure.}
    Non-sink value-state norms are concentrated within each layer-head pair, supporting the approximation \(\|v_i^{(l,h)}\|_2\approx C_{l,h}\) for \(i>0\). 
    }
    \label{fig:ap_value_norms}
\end{figure}

Overall, these diagnostics show that non-sink value norms are sufficiently concentrated to justify a constant-norm approximation, while the sink token should be modeled separately. 

This is the empirical basis for the sink-aware norm assumption used in Theorem~\ref{thm:sink_aware_bound}.

\subsection{Empirical coherence and sink alignment}
\label{ap:value_cosines}

This section supports the coherence assumptions used in the sink-aware deterministic bound. 
For non-sink tokens, the theory uses a worst-case coherence parameter
\[
    |\langle u_p,u_q\rangle| \leq \mu_N,
    \qquad p,q>0,\quad p\neq q.
\]
In experiments, exact maxima are often dominated by outlier pairs. 
Therefore, we report robust empirical versions based on the \(95\)-th percentile of relevant cosine values. 
Specifically, we compute \(\mu_N^{(0.95)}\) from selected-selected and selected-unselected non-sink pairs around the top-\(N\) set.

For the sink token, the theory assumes a signed coherence interval
\[
    -\nu_N^{+}
    \leq
    \langle u_0,u_k\rangle
    \leq
    -\nu_N^{-},
    \qquad k>0.
\]
We evaluate this assumption by measuring the fraction of negative sink-non-sink cosine similarities and robust estimates of \(\nu_N^{-}\) and \(\nu_N^{+}\) from the central \(90\%\) interval of sink-non-sink cosines. 
Concretely, if \(q_{0.05}\) and \(q_{0.95}\) denote the \(5\)-th and \(95\)-th percentiles of \(\langle u_0,u_k\rangle\), then we use
\[
    \nu_N^{+,(0.95)}=\max\{0,-q_{0.05}\},
    \qquad
    \nu_N^{-,(0.95)}=\max\{0,-q_{0.95}\}.
\]

We also report the coefficient of variation of these quantities across samples within each fixed layer-head pair. 
Low CV indicates that \(\mu_N\), \(\nu_N^{-}\), and \(\nu_N^{+}\) behave as stable head-level descriptors rather than sample-specific artifacts.

\begin{figure}[!h]
    \centering
    \begin{subfigure}[h]{0.48\linewidth}
        \centering
        \includegraphics[width=\linewidth]{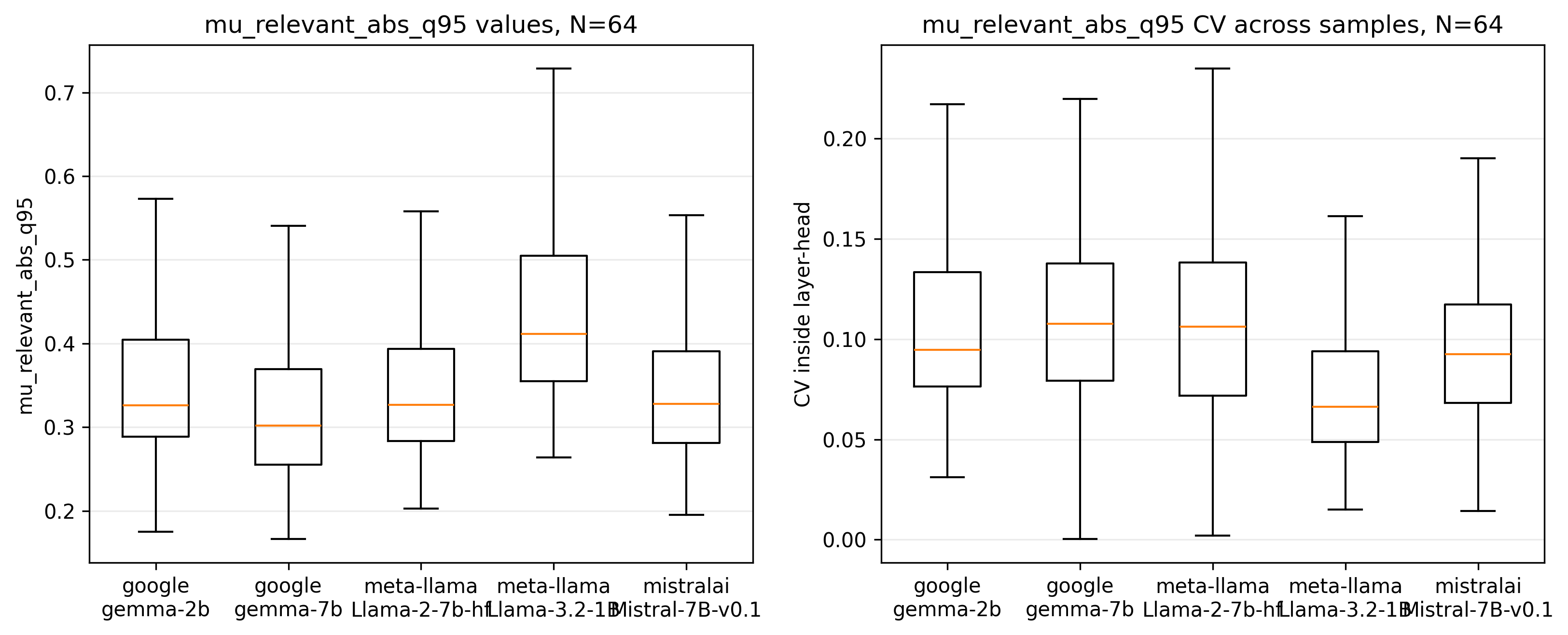}
        \caption{Robust non-sink coherence \(\mu_N^{(0.95)}\).}
        \label{fig:ap_mu_relevant_q95}
    \end{subfigure}
    \hfill
    \begin{subfigure}[h]{0.48\linewidth}
        \centering
        \includegraphics[width=\linewidth]{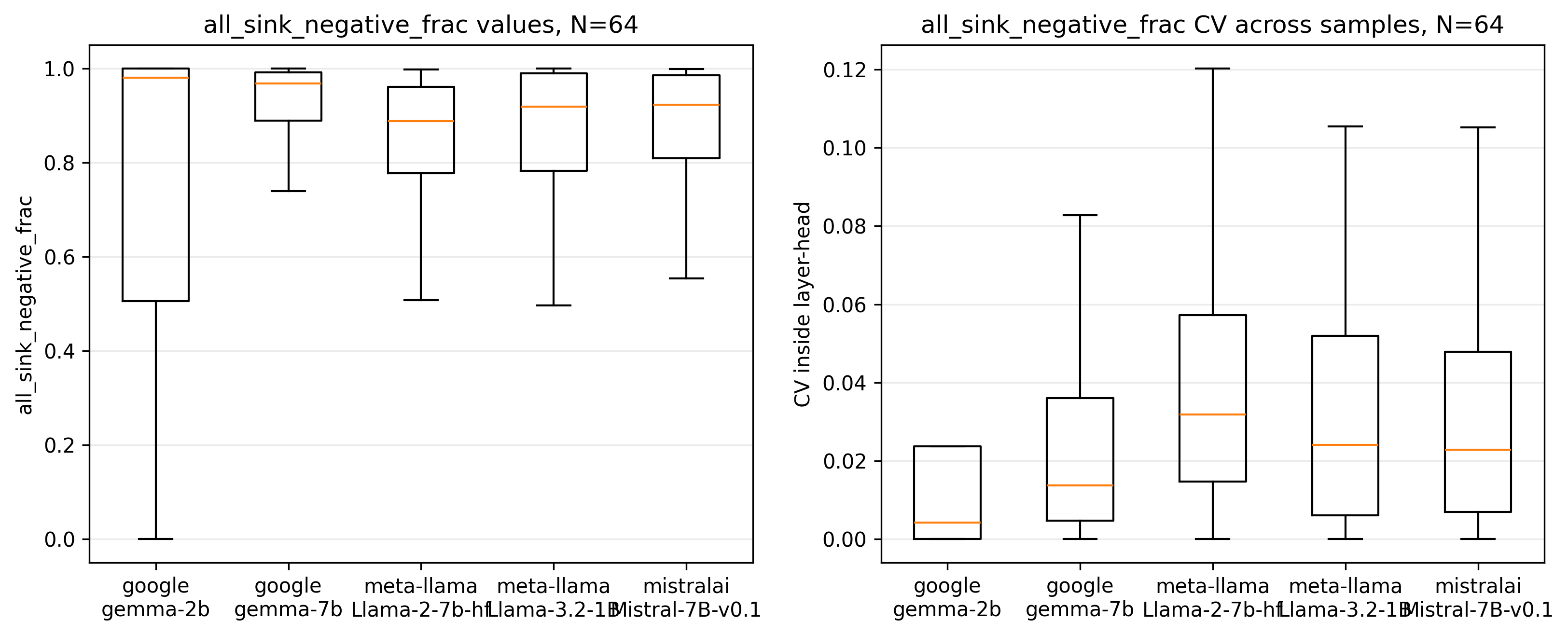}
        \caption{Fraction of negative sink-non-sink cosines.}
        \label{fig:ap_sink_negative_frac}
    \end{subfigure}

    \vspace{0.5em}

    \begin{subfigure}[h]{0.48\linewidth}
        \centering
        \includegraphics[width=\linewidth]{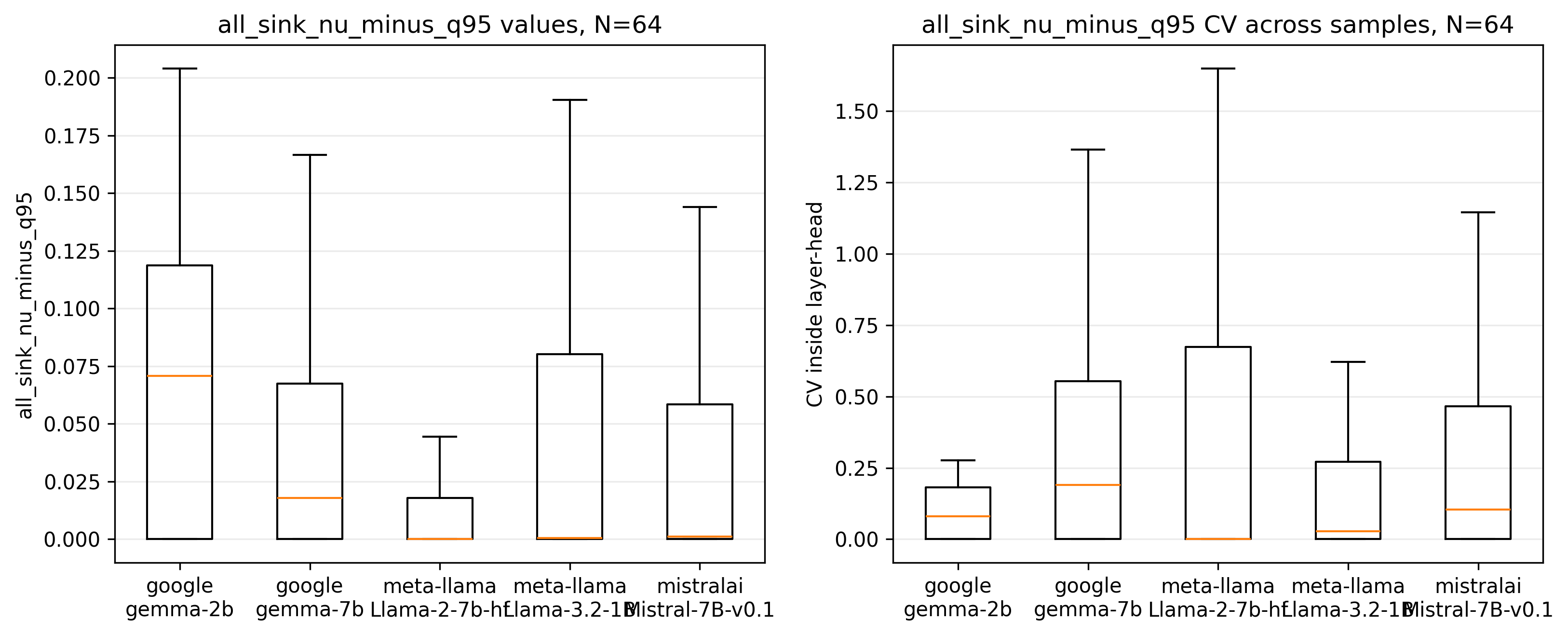}
        \caption{Robust lower sink alignment \(\nu_N^{-,(0.95)}\).}
        \label{fig:ap_nu_minus_q95}
    \end{subfigure}
    \hfill
    \begin{subfigure}[h]{0.48\linewidth}
        \centering
        \includegraphics[width=\linewidth]{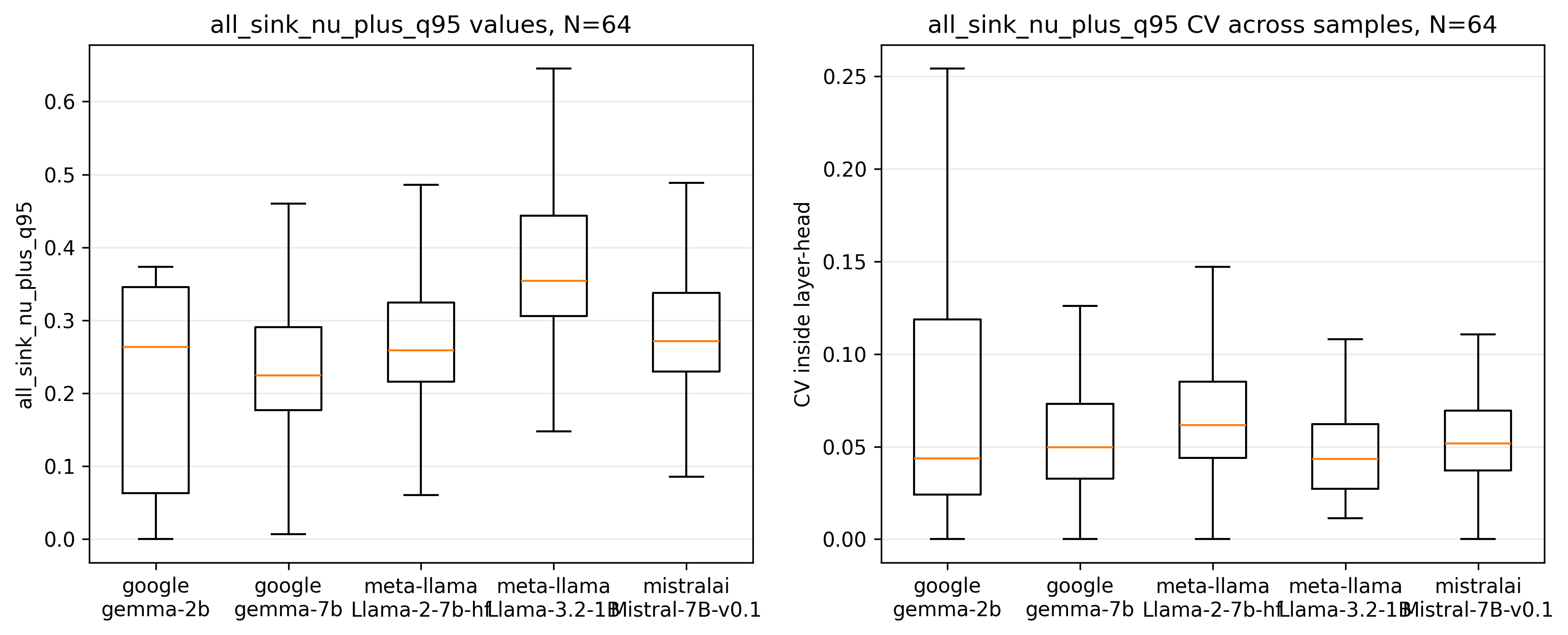}
        \caption{Robust upper sink alignment \(\nu_N^{+,(0.95)}\).}
        \label{fig:ap_nu_plus_q95}
    \end{subfigure}

    \caption{
    \textbf{Empirical coherence parameters.}
    Each panel reports the distribution across layer-head pairs at \(N=64\). 
    For each quantity, the value plot shows its typical magnitude, while the CV plot measures stability across samples inside the same layer-head pair. 
    We use robust quantile estimates because worst-case maxima are highly sensitive to outlier pairs.
    }
    \label{fig:ap_empirical_coherence}
\end{figure}

Figure~\ref{fig:ap_empirical_coherence} supports three conclusions. 
First, non-sink value directions exhibit moderate but stable cosine coherence around top-attention sets. 
Second, sink-non-sink alignments are often negative, supporting the signed sink-coherence assumption. 
Third, the low within-head variability of these quantities indicates that they are meaningful geometric descriptors of heads and layers.

These empirical findings do not make the worst-case theorem tight. 
Rather, they justify the assumptions used to derive a conservative sufficient condition for high precision and recall. 
The robust quantile values describe the typical geometry observed in models, while the theorem itself remains stated in terms of worst-case coherence constants.

\newpage
\subsection{Leave-one-out control}
\label{ap:leave_one_out}

The aggregate $s_N$ contains the selected contributions themselves, so selected tokens may appear close to $s_N$ partly because of self-inclusion. 
We therefore use a leave-one-out control. 
For each selected token $i\in I_N$, define
\begin{equation}
    s_{N,-i}=\sum_{k\in I_N\setminus\{i\}}y_k .
\end{equation}
We measure whether selected contributions remain aligned with the selected set after removing their own contribution:
\begin{equation}
    C_N^{\mathrm{loo}}
    =
    \frac{1}{N}
    \sum_{i\in I_N}
    \cos(y_i,s_{N,-i}).
\end{equation}
We also define the contrastive leave-one-out margin
\begin{equation}
    M_N^{\mathrm{loo}}
    =
    \frac{1}{N}
    \sum_{i\in I_N}
    \cos(y_i,s_{N,-i})
    -
    \frac{1}{L-N}
    \sum_{j\notin I_N}
    \cos(y_j,s_N).
\end{equation}
Positive values indicate that selected contributions are more mutually coherent than unselected contributions, even after removing self-inclusion.

\begin{figure*}[!h]
    \centering
    \includegraphics[width=0.48\linewidth]{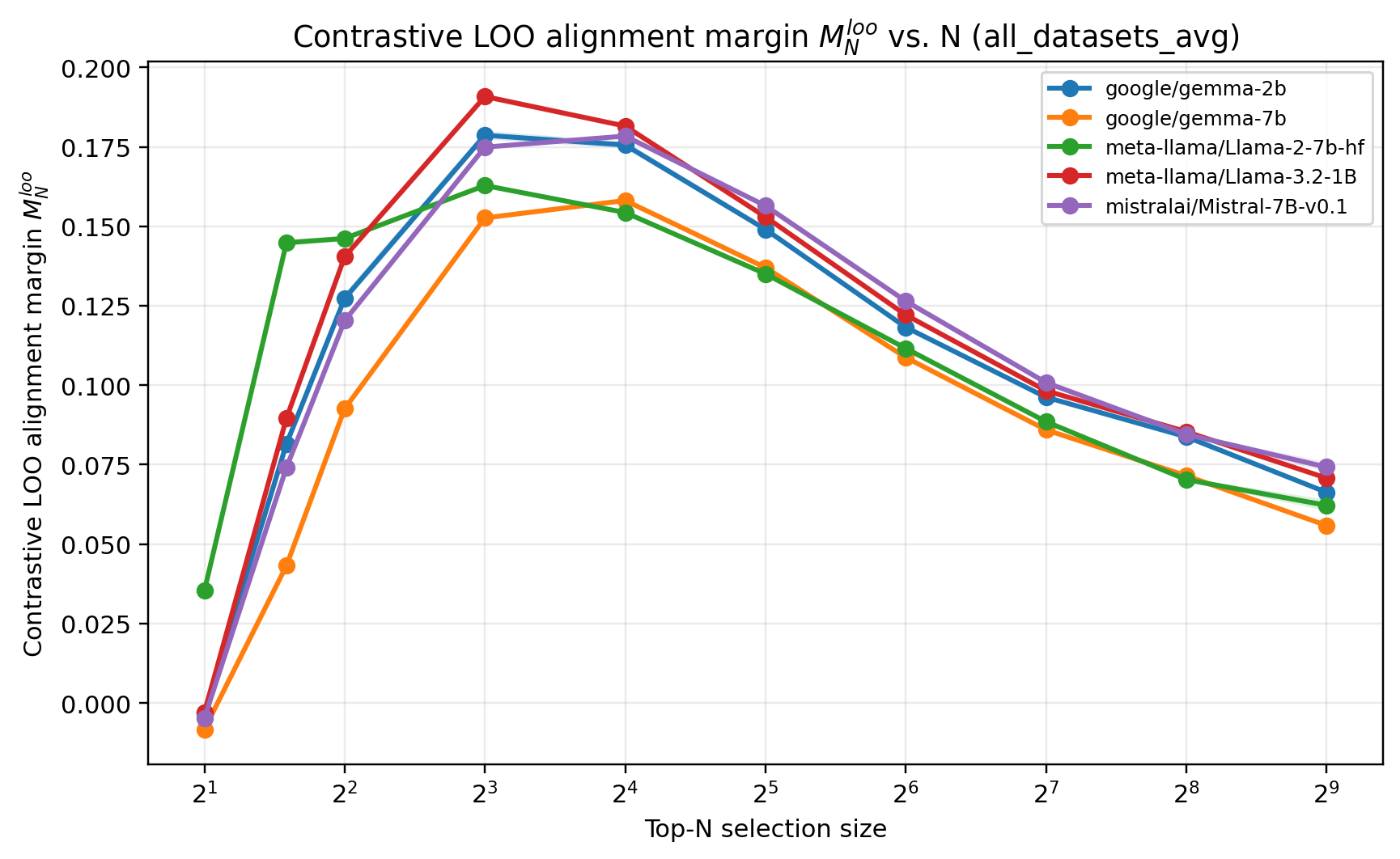}
    \includegraphics[width=0.48\linewidth]{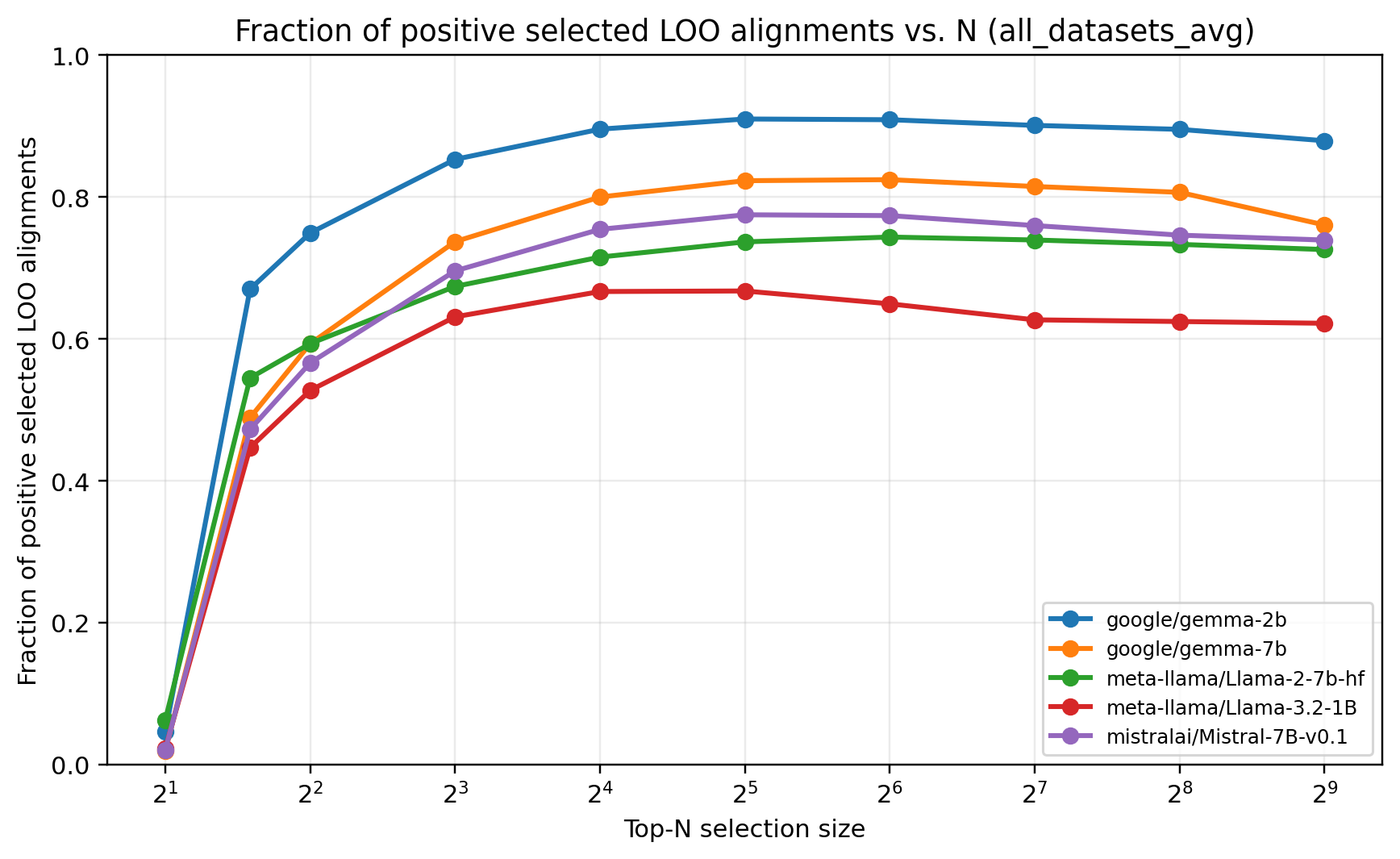}
    \caption{
    \textbf{Leave-one-out robustness of top-$N$ geometric separability.}
    \textbf{Left:} Contrastive leave-one-out alignment margin
    $M_N^{\mathrm{loo}}$, comparing selected leave-one-out alignment against
    unselected alignment to the full selected aggregate.
    Positive values indicate that selected contributions remain more mutually
    coherent than unselected contributions, even after removing self-inclusion.
    \textbf{Right:} Fraction of selected contributions with positive
    leave-one-out alignment. Across model families, the small-to-moderate
    $N$ regime exhibits consistently positive leave-one-out structure.
    }
    \label{fig:loo_robustness}
\end{figure*}

We also evaluate a stricter distance-based leave-one-out margin,
\begin{equation}
    \gamma_N^{\mathrm{loo}}
    =
    \min_{j\notin I_N}\|y_j-s_N\|_2
    -
    \max_{i\in I_N}\|y_i-s_{N,-i}\|_2 .
\end{equation}
This quantity is a worst-case test and is often negative for larger $N$, indicating that strict Euclidean separation is stronger than the directional coherence needed for the classifier view.

\begin{figure}[!h]
    \centering
    \includegraphics[width=0.8\linewidth]{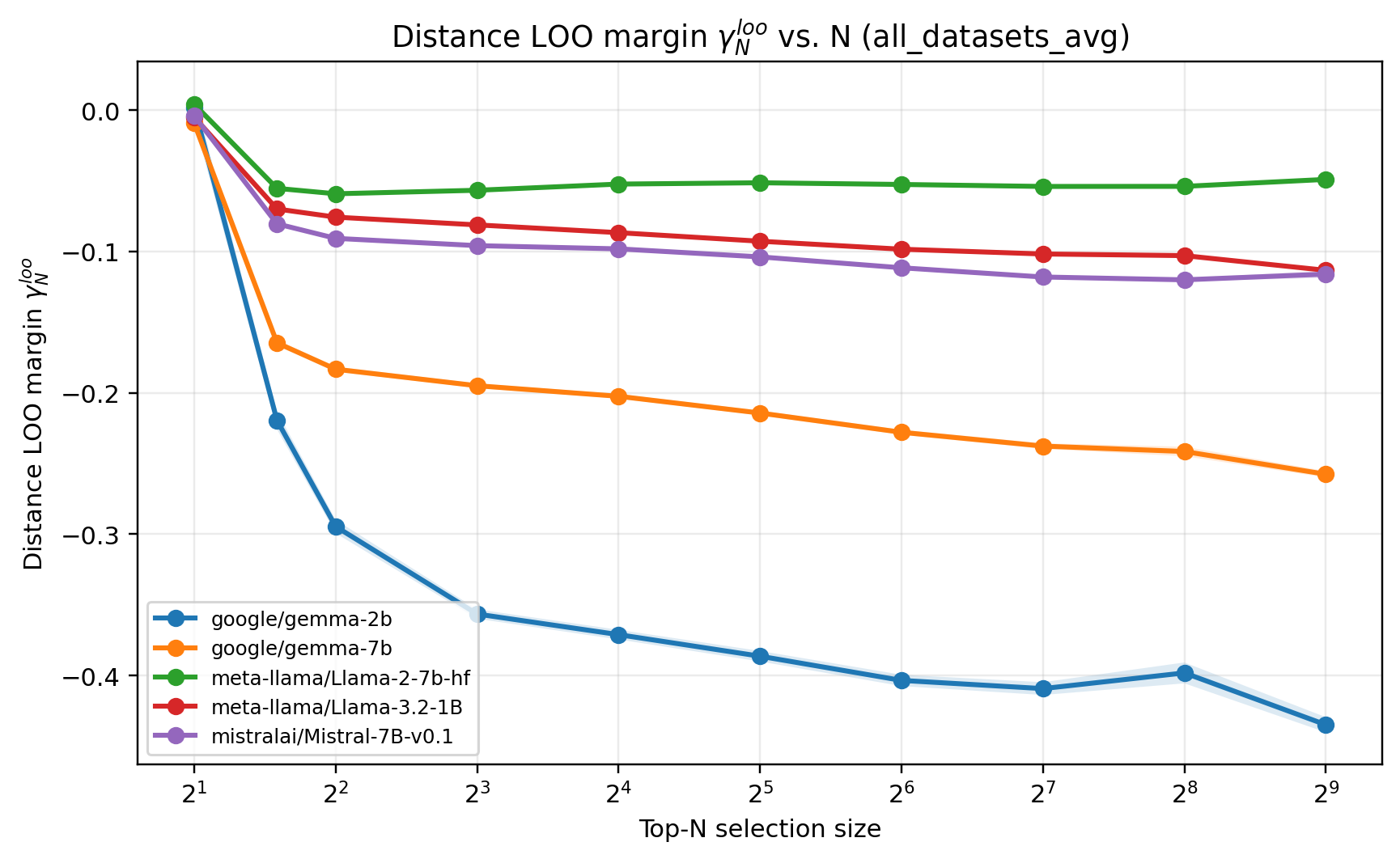}
    \caption{
    \textbf{Strict distance-based leave-one-out separability margin.}
    We evaluate the worst-case leave-one-out distance margin
    $\gamma_N^{\mathrm{loo}}=
    \min_{j\notin I_N}\|y_j-s_N\|_2
    -
    \max_{i\in I_N}\|y_i-s_{N,-i}\|_2$.
    Positive values would imply that every selected contribution remains closer
    to the leave-one-out selected aggregate than every unselected contribution
    is to the full selected aggregate. The margin is close to zero only at very
    small $N$ and becomes negative as $N$ grows, indicating that strict
    worst-case distance separation is generally too strong for attention-scaled
    contribution geometry.
    }
    \label{fig:distance_loo_margin}
\end{figure}

Figure~\ref{fig:distance_loo_margin} shows that the distance-based
leave-one-out margin is mostly negative beyond the smallest selection sizes.
This means that top-attended contributions do not generally form a perfectly
distance-separated cluster in the worst-case sense: for most $N$, at least one
unselected contribution can be closer to the selected aggregate than the
least-compatible selected contribution is to its leave-one-out aggregate.
\textbf{This does not contradict the positive contrastive alignment results.} Instead,
it shows that the appropriate LOO conclusion is directional coherence rather
than strict Euclidean separation.

\subsection{Long-context precision and recall}
\label{ap:long_context_precision_recall}

We additionally evaluate the extremal precision and recall at different context lengths. 
For each model and selection size \(N\), we compute \(P(r_{\max},N)\) and \(R(r_{\min},N)\) at \(L\in\{512,1024,2048\}\), averaging over datasets, layers, heads, and samples. 

\begin{figure}[!h]
    \centering
    \includegraphics[width=0.9\linewidth]{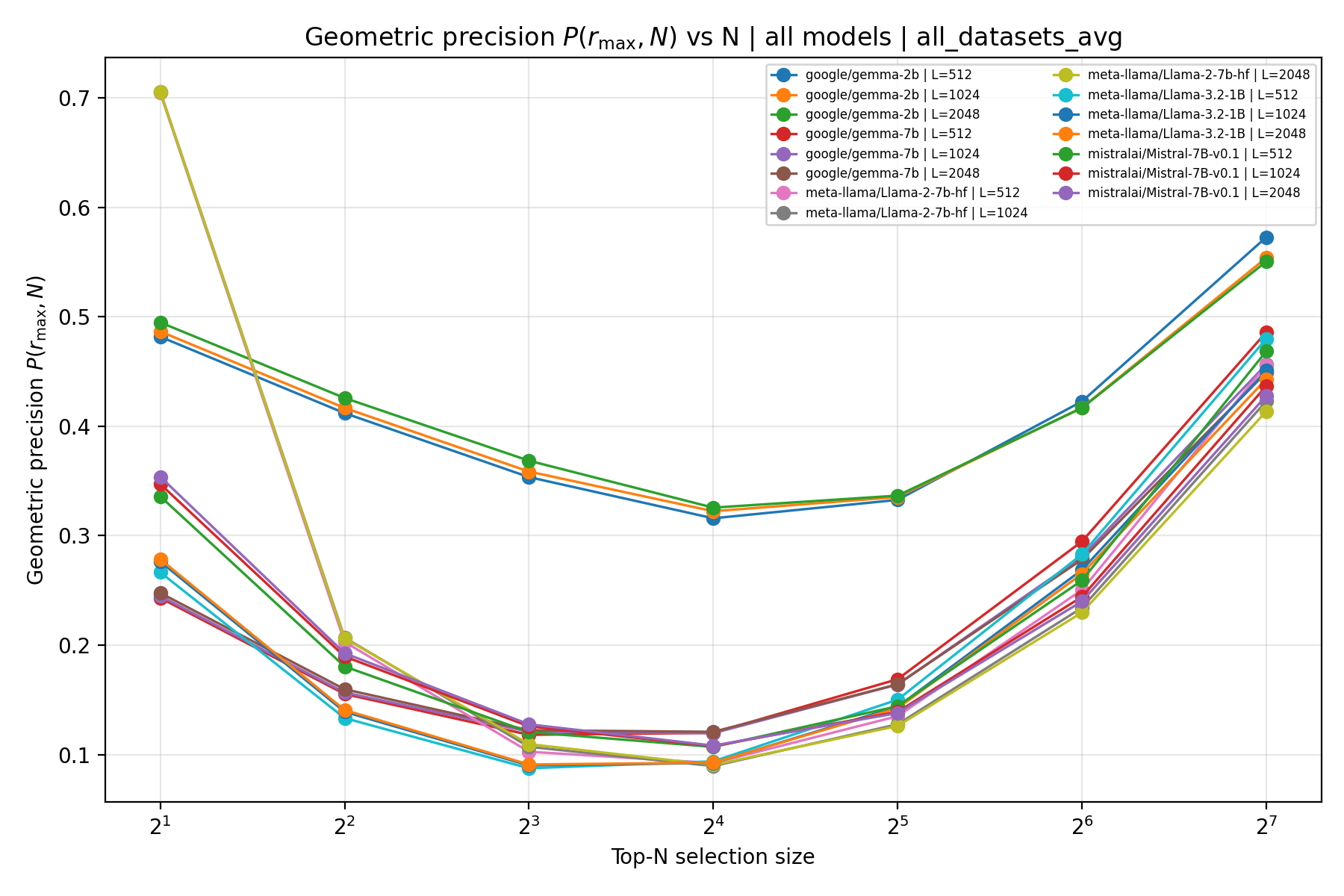}
    \caption{
    \textbf{Long-context precision.}
    Geometric precision \(P(r_{\max},N)\) across context lengths \(L\in\{512,1024,2048\}\). 
    Precision is highest in the sparse regime, drops at intermediate \(N\), and partially recovers at larger \(N\). 
    }
    \label{fig:ap_long_context_precision}
\end{figure}

\begin{figure}[!h]
    \centering
    \includegraphics[width=0.9\linewidth]{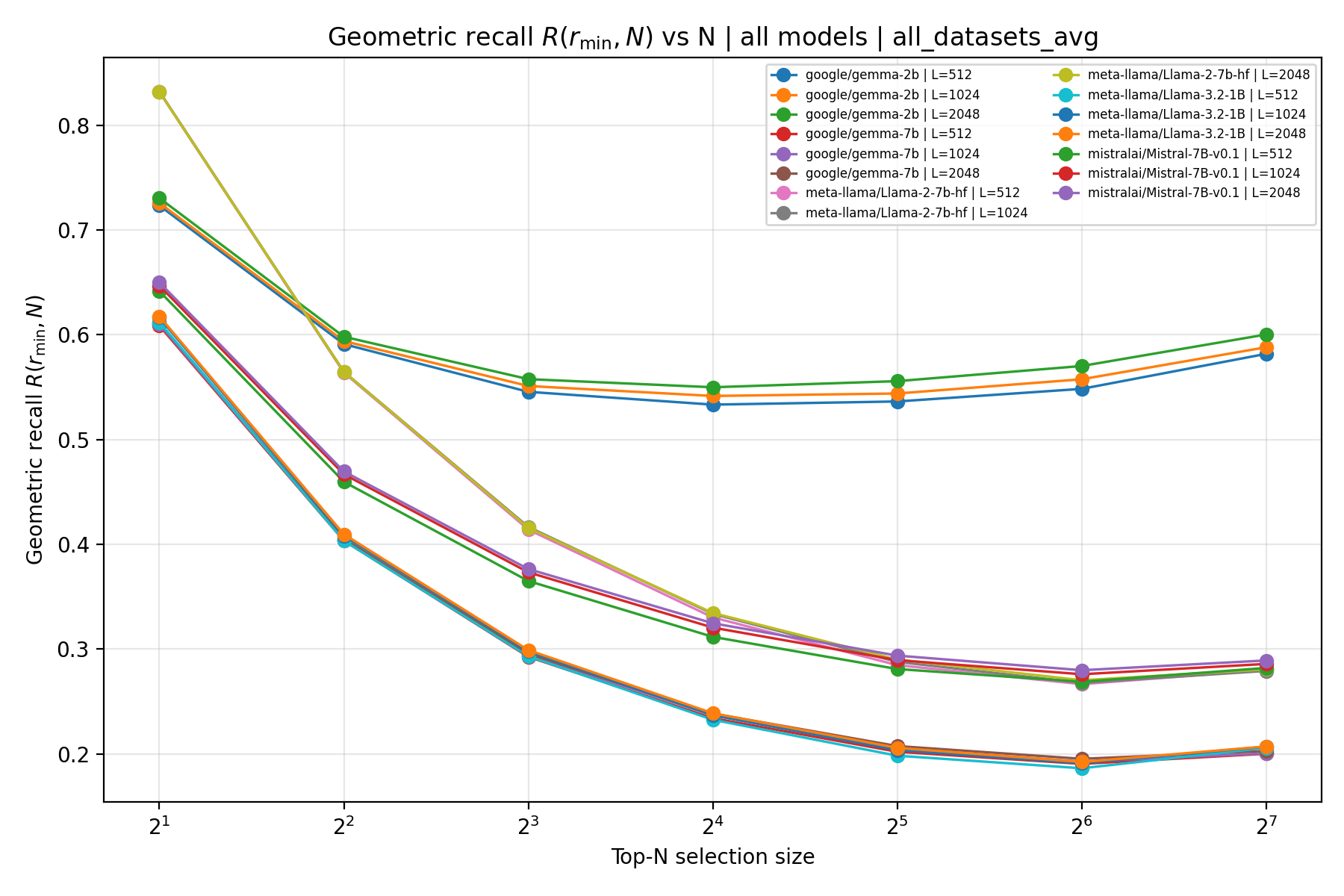}
    \caption{
    \textbf{Long-context recall.}
    Geometric recall \(R(r_{\min},N)\) across context lengths \(L\in\{512,1024,2048\}\). 
    Recall decreases as \(N\) grows from the sparse regime, indicating that selected tokens become less uniformly close to the aggregate. 
    }
    \label{fig:ap_long_context_recall}
\end{figure}

\subsection{Dataset-specific precision and recall}
\label{ap:dataset_specific_precision_recall}

We additionally report geometric precision and recall separately for WikiText-103 and OpenWebText. 
For each dataset, we average across models, layers, heads, samples, and context lengths \(L\in\{512,1024,2048\}\), while keeping model-specific curves separate. 
The goal is to verify that the main geometric patterns are not tied to a particular text source.

Across both datasets, the overall behavior is consistent. 
Geometric recall \(R(r_{\min},N)\) is highest at small \(N\) and decreases as the selection size grows, showing that small top-\(N\) sets are more tightly concentrated around their aggregate. 
Geometric precision \(P(r_{\max},N)\) also changes systematically with \(N\), with similar curve shapes across WikiText-103 and OpenWebText. 
While absolute values vary across models and context lengths, the qualitative trends remain stable across datasets.

These results support the claim that the geometric classifier view is robust to dataset choice: the selected top-attention sets exhibit similar precision-recall structure on both curated text (WikiText-103) and broader web text (OpenWebText).

\begin{figure}[!h]
    \centering
    \begin{subfigure}[!h]{0.48\linewidth}
        \centering
        \includegraphics[width=\linewidth]{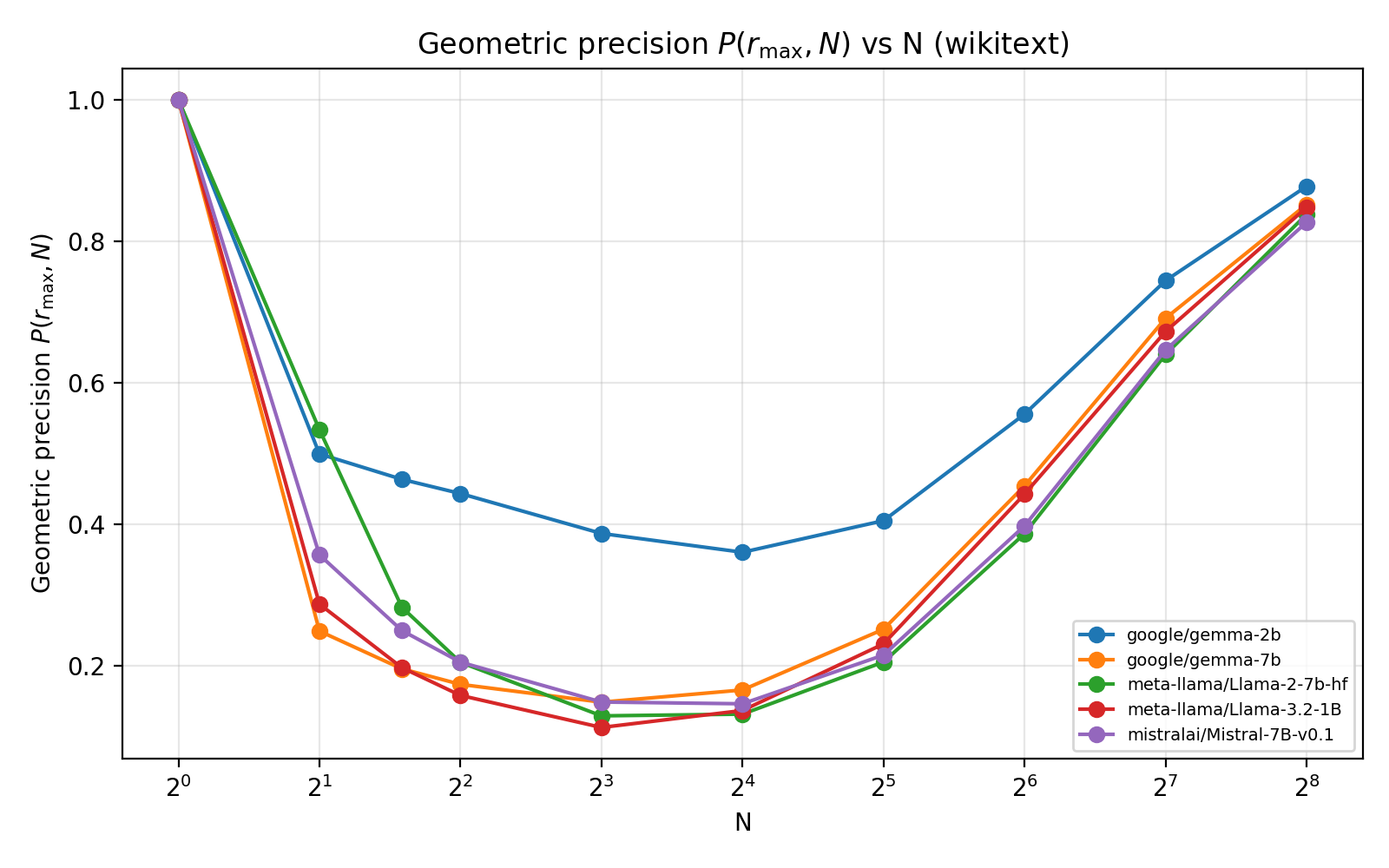}
        \caption{WikiText-103: geometric precision \(P(r_{\max},N)\).}
        \label{fig:ap_precision_wikitext}
    \end{subfigure}
    \hfill
    \begin{subfigure}[!h]{0.48\linewidth}
        \centering
        \includegraphics[width=\linewidth]{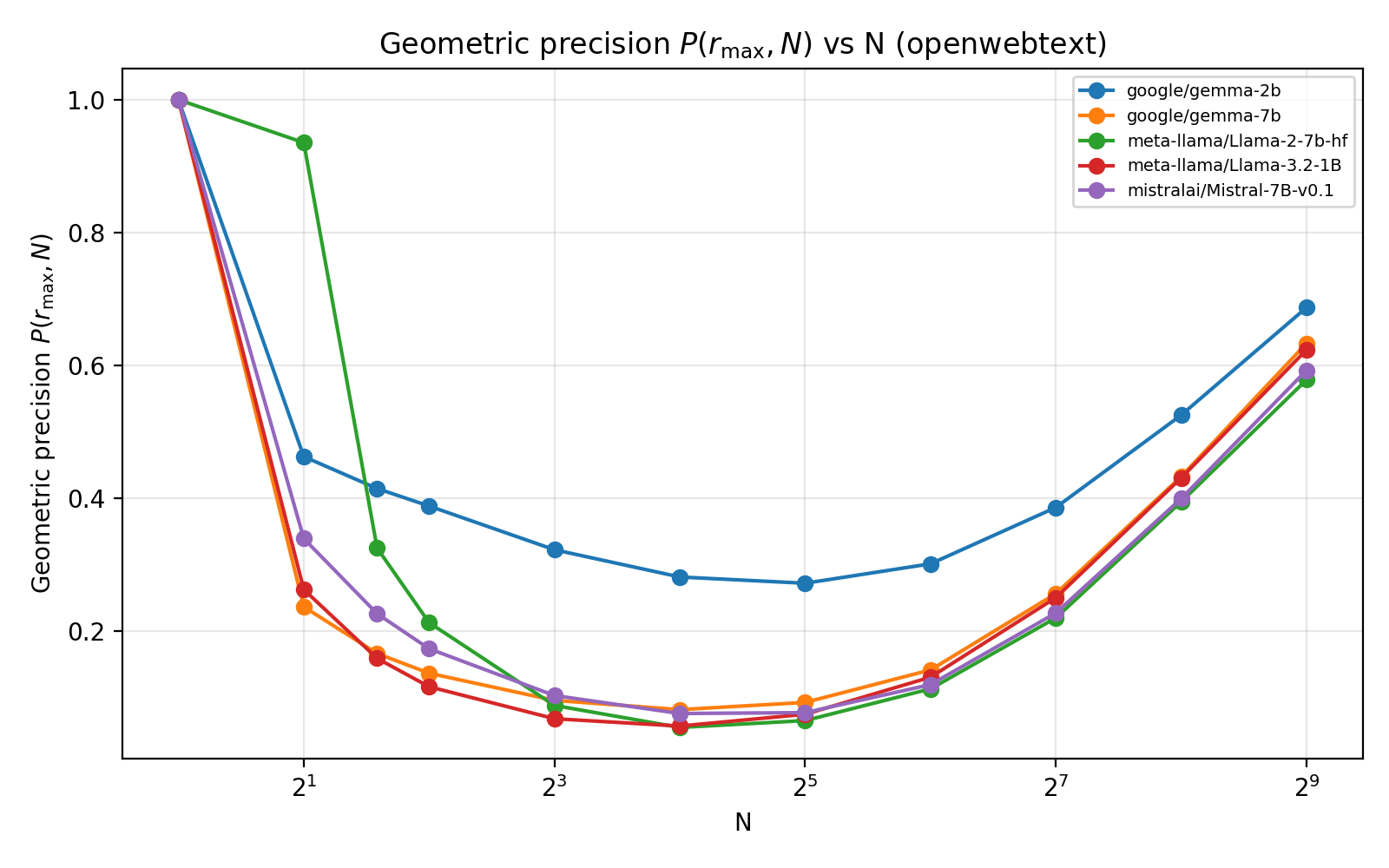}
        \caption{OpenWebText: geometric precision \(P(r_{\max},N)\).}
        \label{fig:ap_precision_openwebtext}
    \end{subfigure}

    \vspace{0.5em}

    \begin{subfigure}[!h]{0.48\linewidth}
        \centering
        \includegraphics[width=\linewidth]{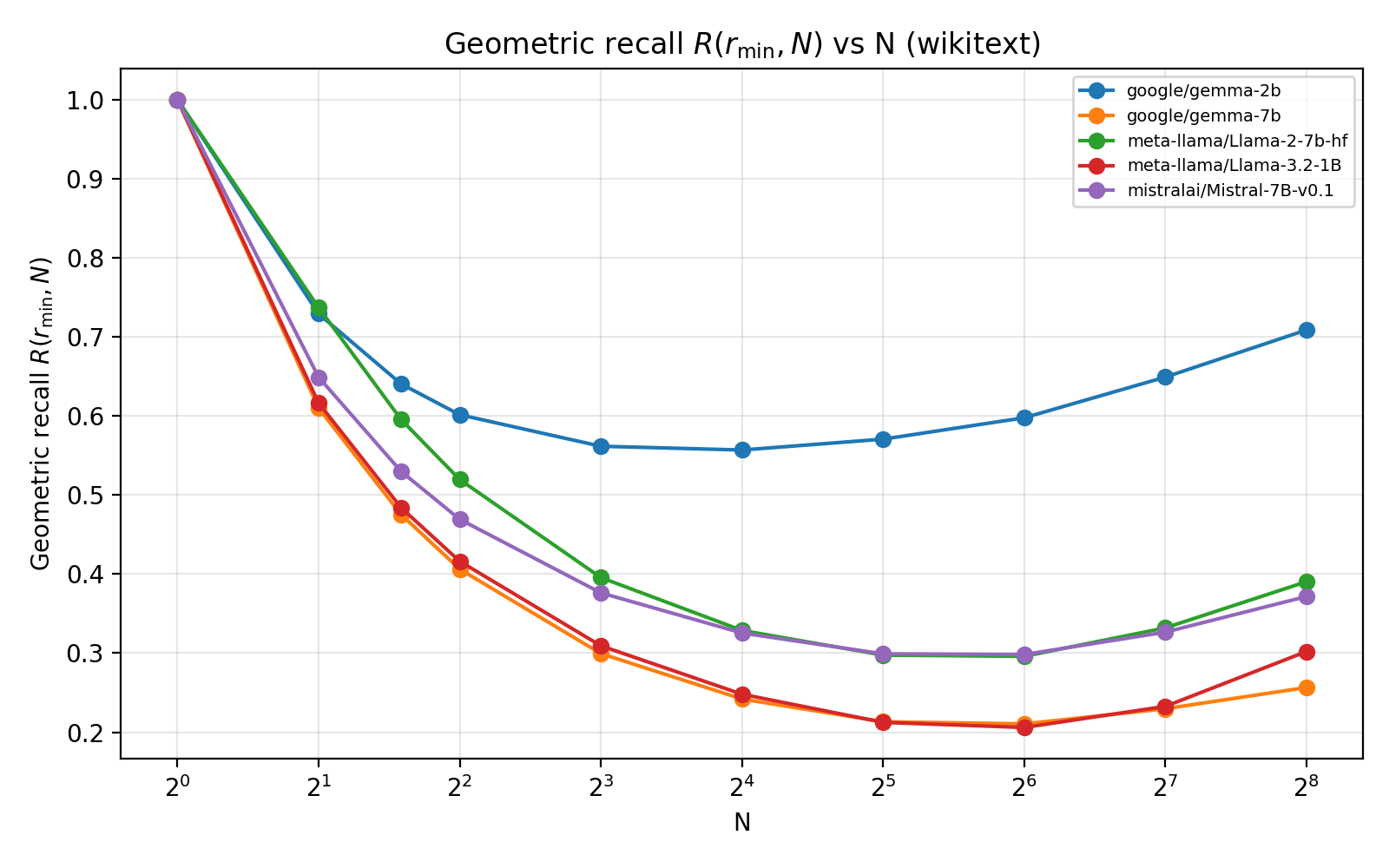}
        \caption{WikiText-103: geometric recall \(R(r_{\min},N)\).}
        \label{fig:ap_recall_wikitext}
    \end{subfigure}
    \hfill
    \begin{subfigure}[!h]{0.48\linewidth}
        \centering
        \includegraphics[width=\linewidth]{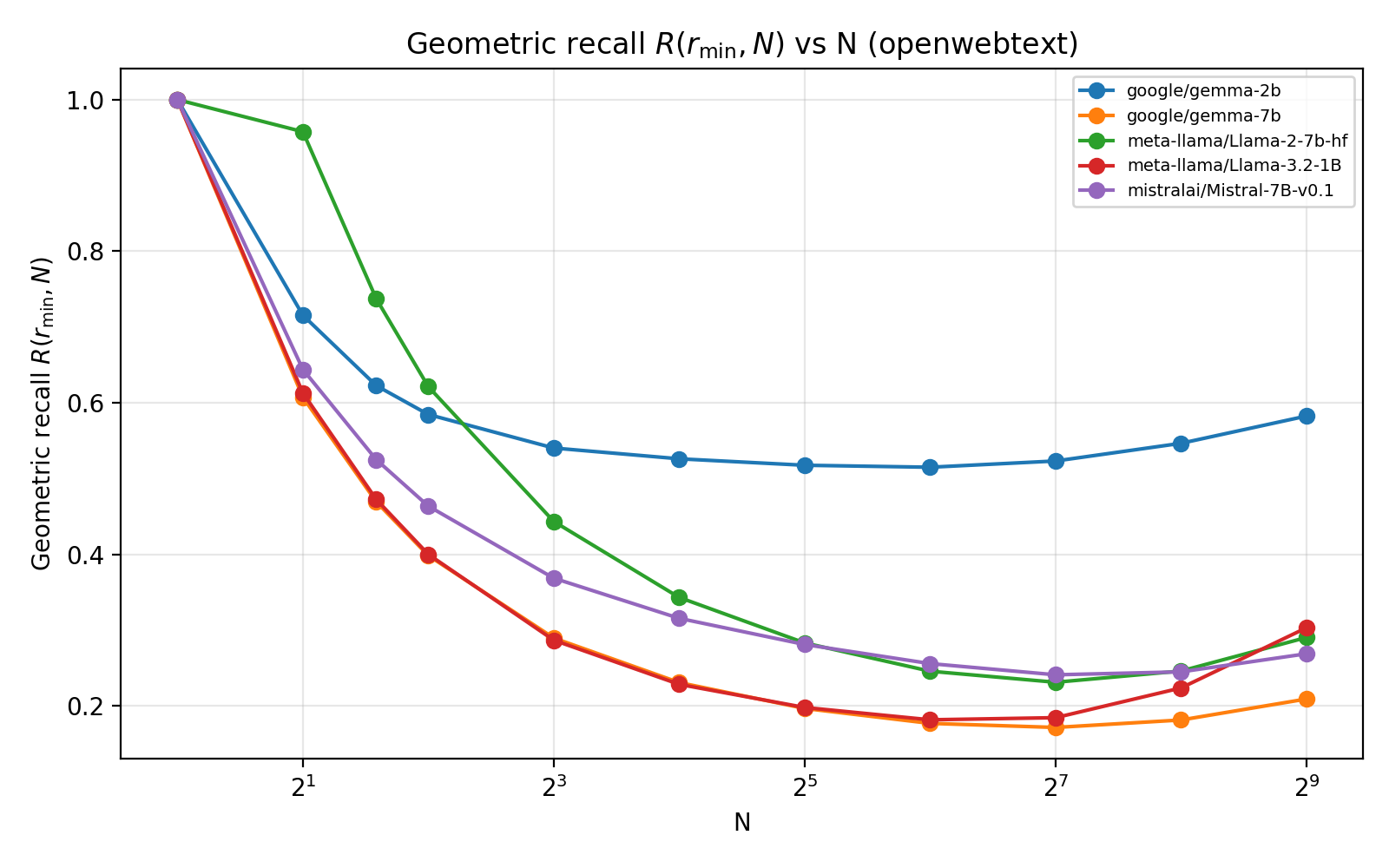}
        \caption{OpenWebText: geometric recall \(R(r_{\min},N)\).}
        \label{fig:ap_recall_openwebtext}
    \end{subfigure}

    \caption{
    Dataset-specific geometric precision and recall across models and context lengths. 
    The qualitative dependence on \(N\) is similar on WikiText-103 and OpenWebText: small top-\(N\) sets are more geometrically coherent, while larger \(N\) leads to weaker recall and changing neighborhood purity. 
    This shows that the observed geometric classifier structure is stable across different text domains.
    }
    \label{fig:ap_dataset_specific_precision_recall}
\end{figure}

\newpage

\subsection{Additional attention sink diagnostics}
\label{ap:sink_effect}

This appendix gives additional diagnostics for the attention sink. 
The main text shows that the sink has a smaller norm, a distinct directional alignment, and a measurable effect on the geometric separability score. 
Here we further decompose this effect into selection frequency, aggregate shift, directional change, and sink-related inversions.

\begin{figure}[h]
    \centering
    \includegraphics[width=\linewidth]{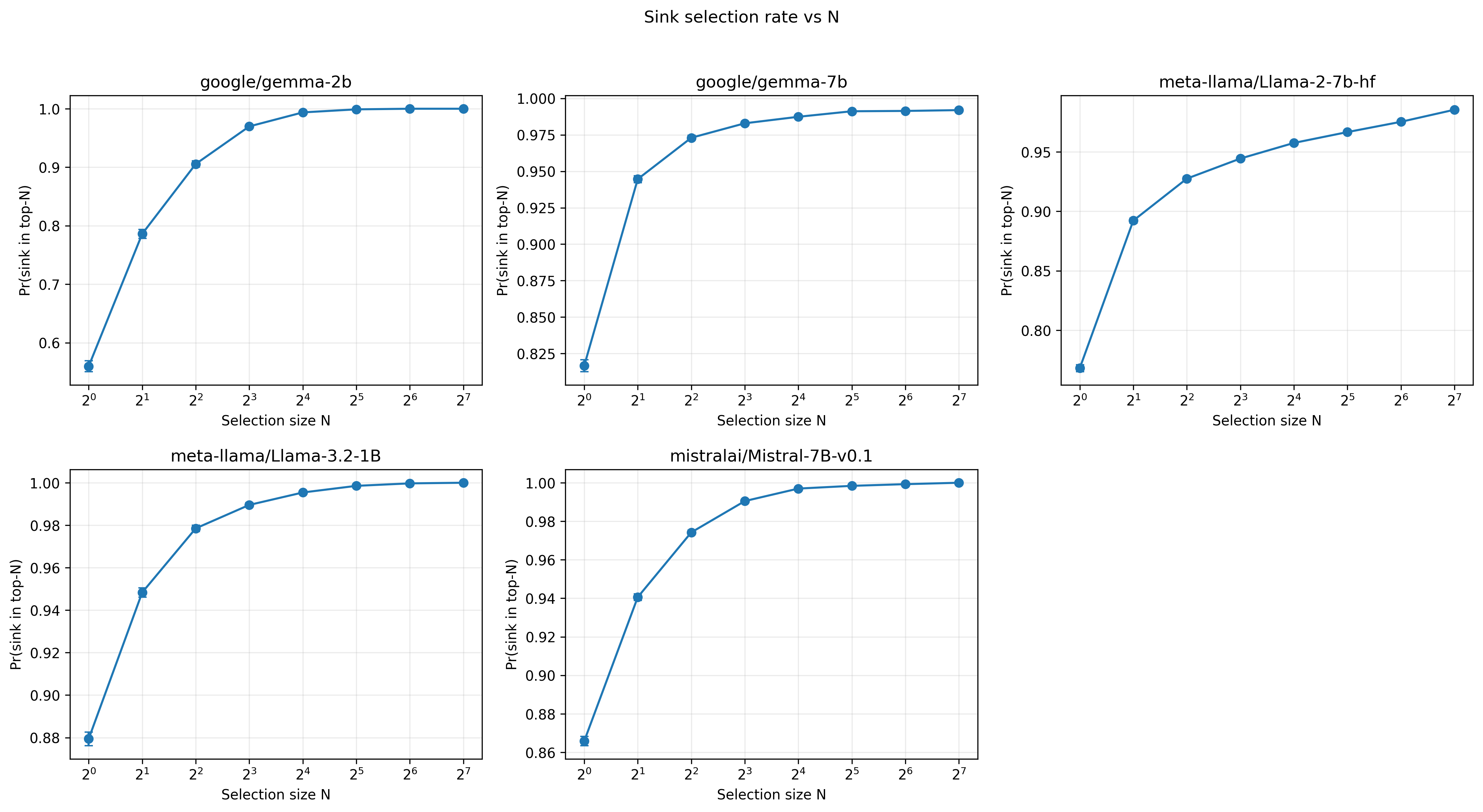}
    \caption{
    \textbf{Sink selection rate.}
    The first token is frequently included in the top-\(N\) attention set. 
    This explains why sink geometry can strongly affect the selected aggregate \(s_N\).
    }
    \label{fig:ap_sink_selection_rate}
\end{figure}

First, we measure how often the sink belongs to the top-\(N\) attention set. 
Figure~\ref{fig:ap_sink_selection_rate} shows that the sink is selected very frequently, even for small \(N\). 
Thus, sink geometry is not a rare edge case: it directly affects the top-attention selected set in most layer-head instances.

We also condition the geometric separability score on whether the sink is selected. 
This diagnostic is useful but should be interpreted carefully, since the two groups are not matched: heads and samples where the sink is selected may differ systematically from those where it is not selected.

\begin{figure}[h]
    \centering
    \includegraphics[width=\linewidth]{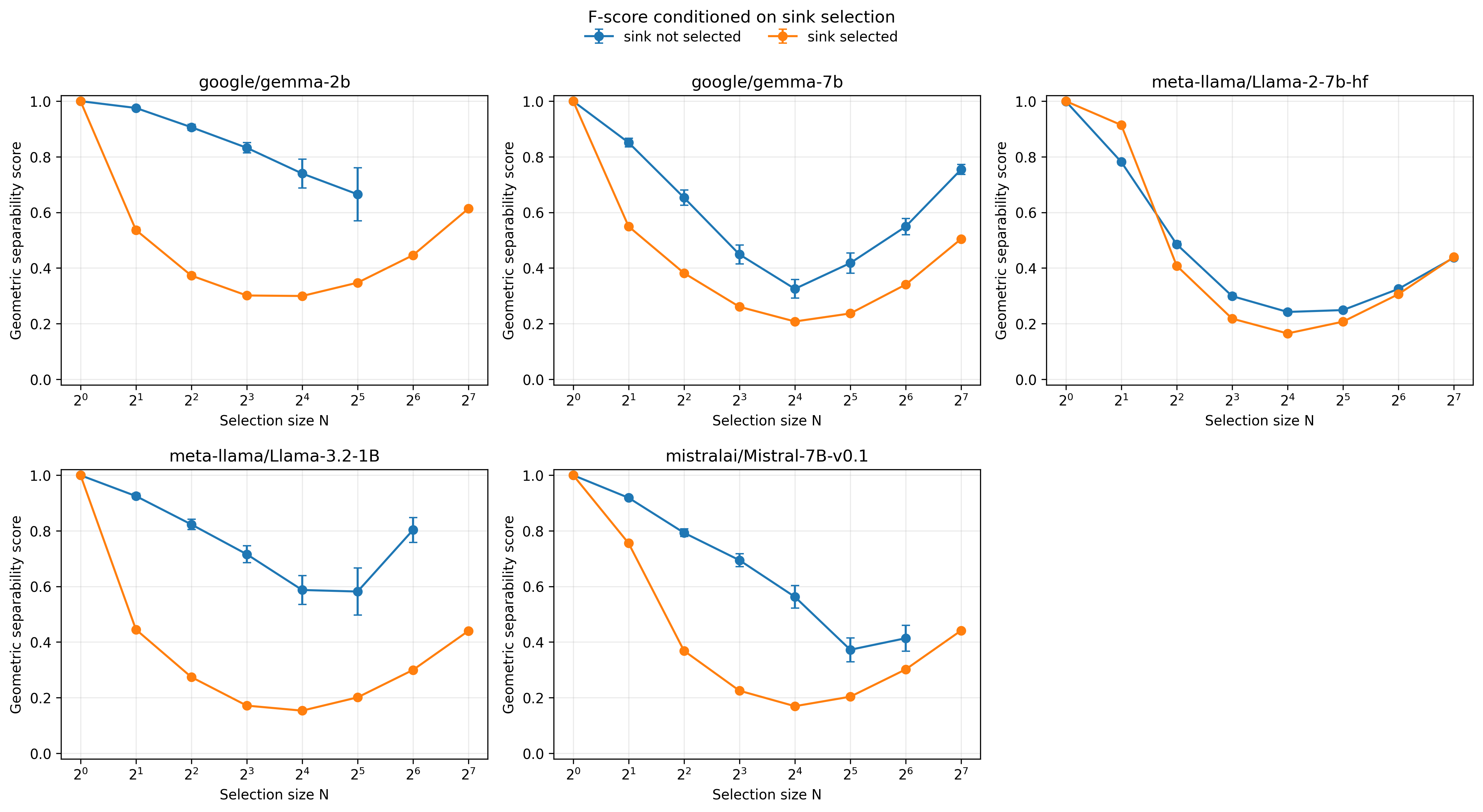}
    \caption{
    \textbf{Separability conditioned on sink selection.}
    The geometric separability score differs depending on whether the sink is included in \(I_N\). 
    This supports the view that sink selection changes the geometry of the selected class.
    }
    \label{fig:ap_sink_conditioned_gscore}
\end{figure}

\begin{figure}[h]
    \centering
    \includegraphics[width=\linewidth]{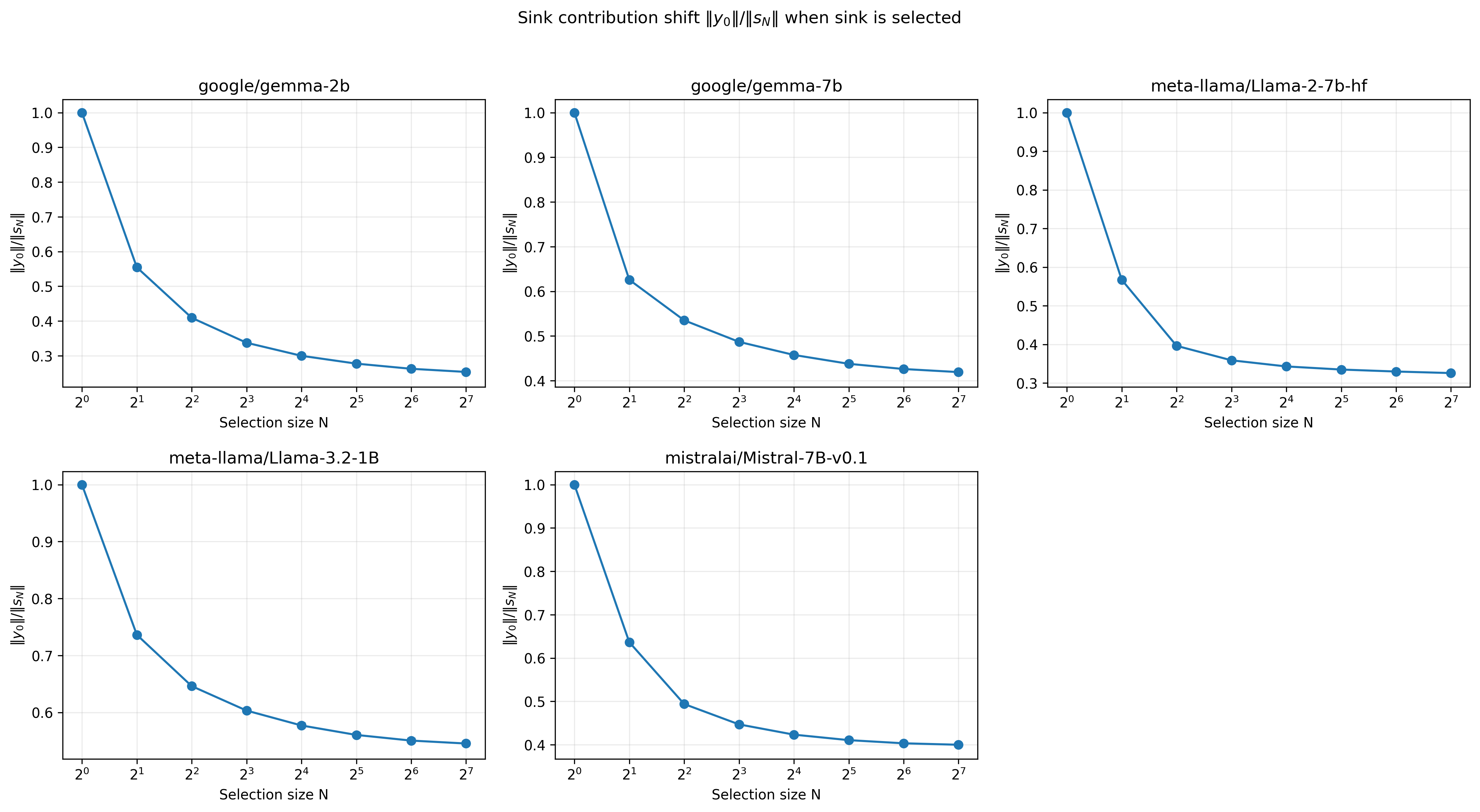}
    \caption{
    \textbf{Sink contribution magnitude.}
    We report \(\|y_0\|_2/\|s_N\|_2\) when the sink is selected. 
    The sink contribution is largest for small \(N\), where it can dominate the selected aggregate.
    }
    \label{fig:ap_sink_shift}
\end{figure}

\begin{figure}[h]
    \centering
    \includegraphics[width=\linewidth]{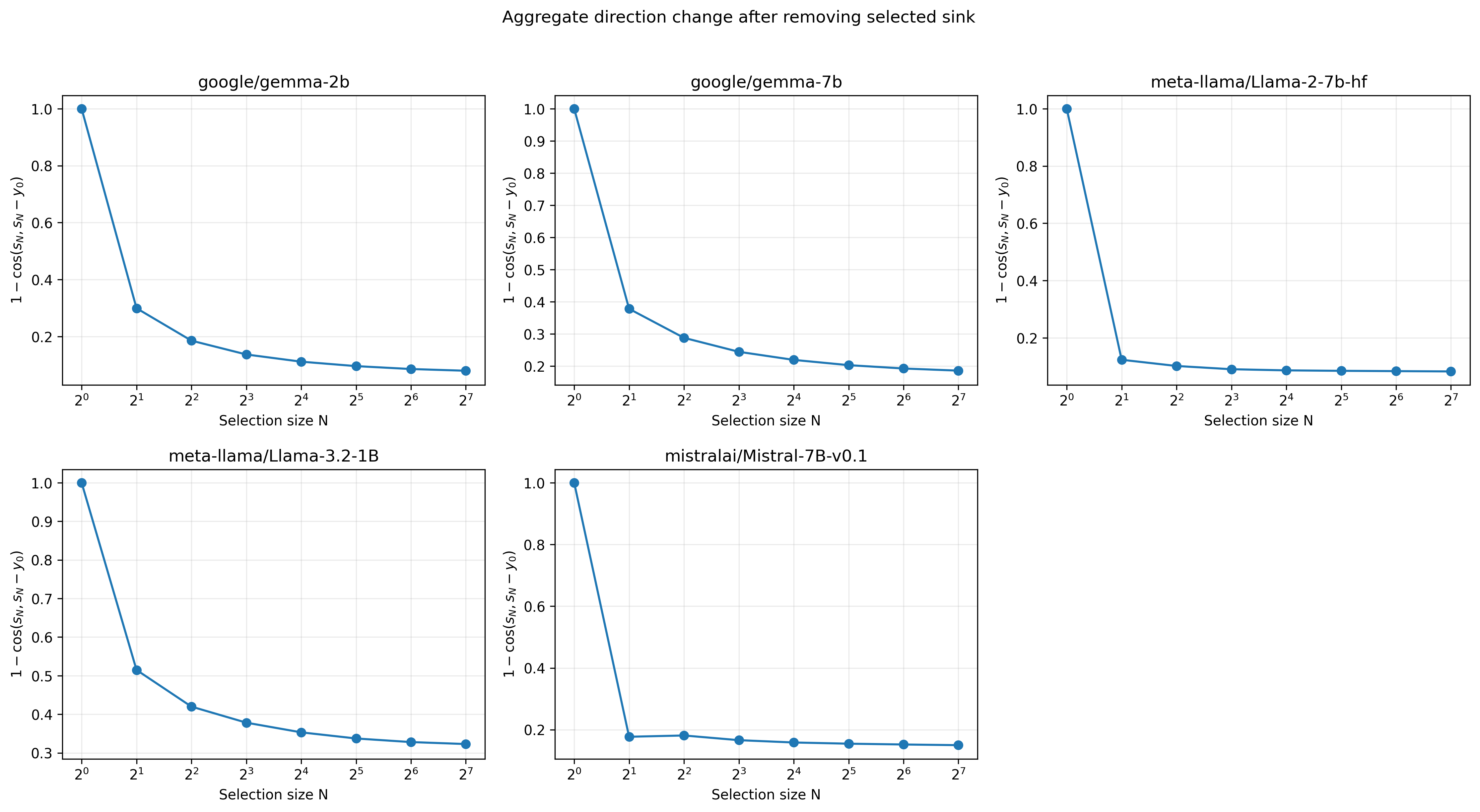}
    \caption{
    \textbf{Aggregate direction change after removing the sink.}
    We report \(1-\cos(s_N,s_N-y_0)\) when the sink is selected. 
    The sink can strongly affect the direction of the selected aggregate, especially in sparse top-\(N\) regimes.
    }
    \label{fig:ap_sink_direction_change}
\end{figure}

When the sink is selected, we quantify how strongly it contributes to the aggregate by measuring
\[
    \frac{\|y_0\|_2}{\|s_N\|_2}.
\]
We also measure the directional change caused by removing the selected sink,
\[
    1-\cos(s_N,s_N-y_0).
\]
Figures~\ref{fig:ap_sink_shift} and~\ref{fig:ap_sink_direction_change} show that the sink can account for a large fraction of the aggregate at small \(N\), and that removing it can substantially change the aggregate direction. 
The effect decreases as \(N\) grows, because the aggregate contains more non-sink contributions.

Next, we study the case where the sink is not selected. 
An unselected sink can still affect precision if it lies inside the selected ball. 
We therefore measure the intrusion event
\[
    D_0 \leq r_{\max}^2,
\]
where \(D_i=\|y_i-s_N\|_2^2\). 
We also measure the fraction of selected tokens beaten by the unselected sink,
\[
    \frac{1}{N}\sum_{i\in I_N}\mathbf{1}\{D_0\leq D_i\}.
\]
These quantities correspond to sink-related selected-unselected inversions.

\begin{figure}[h]
    \centering
    \includegraphics[width=\linewidth]{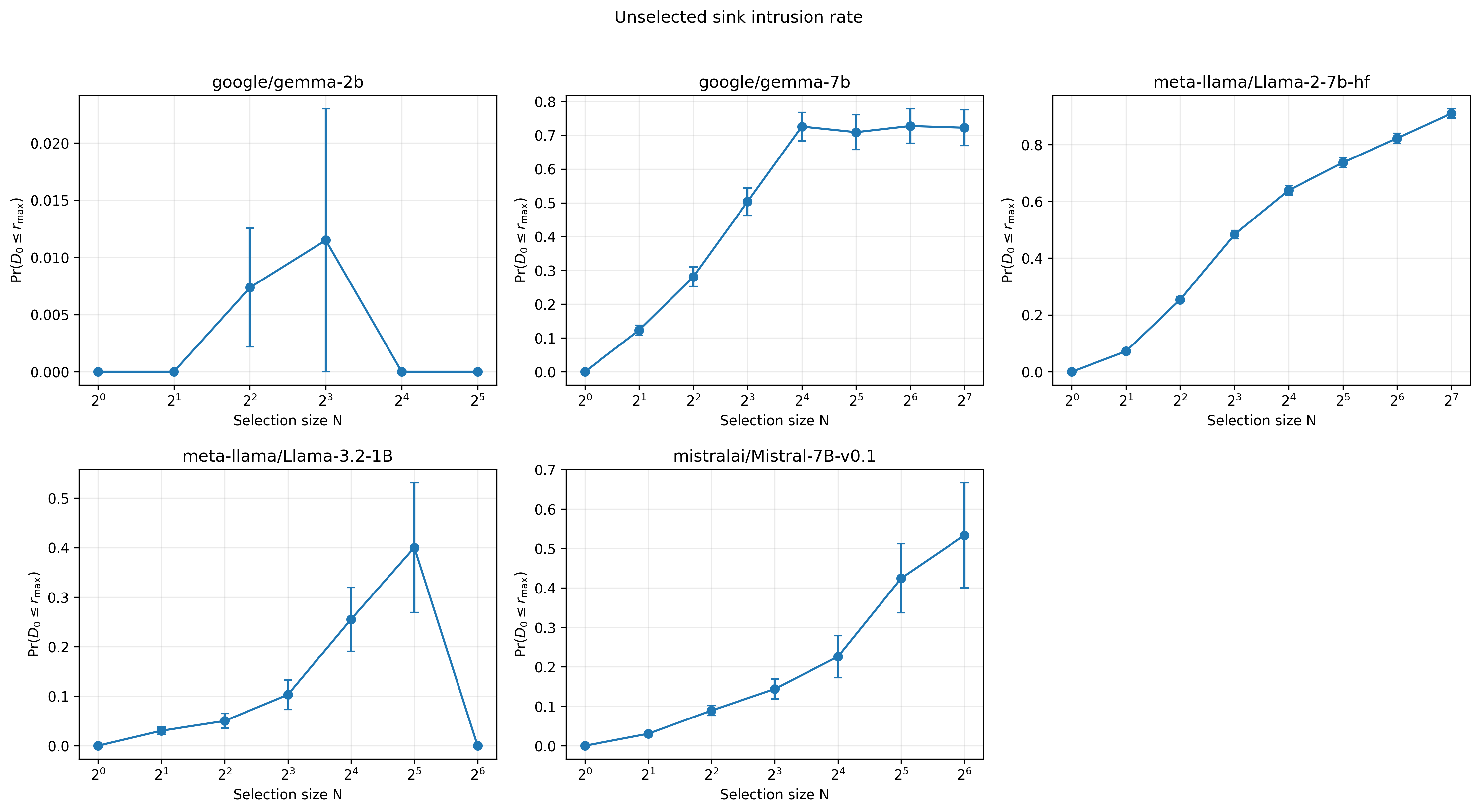}
    \caption{
    \textbf{Unselected sink intrusion rate.}
    We report the probability that the unselected sink lies inside the selected ball, \(D_0\leq r_{\max}^2\). 
    This measures whether the sink creates false positives for the geometric classifier.
    }
    \label{fig:ap_sink_intrusion}
\end{figure}

\begin{figure}[h]
    \centering
    \includegraphics[width=\linewidth]{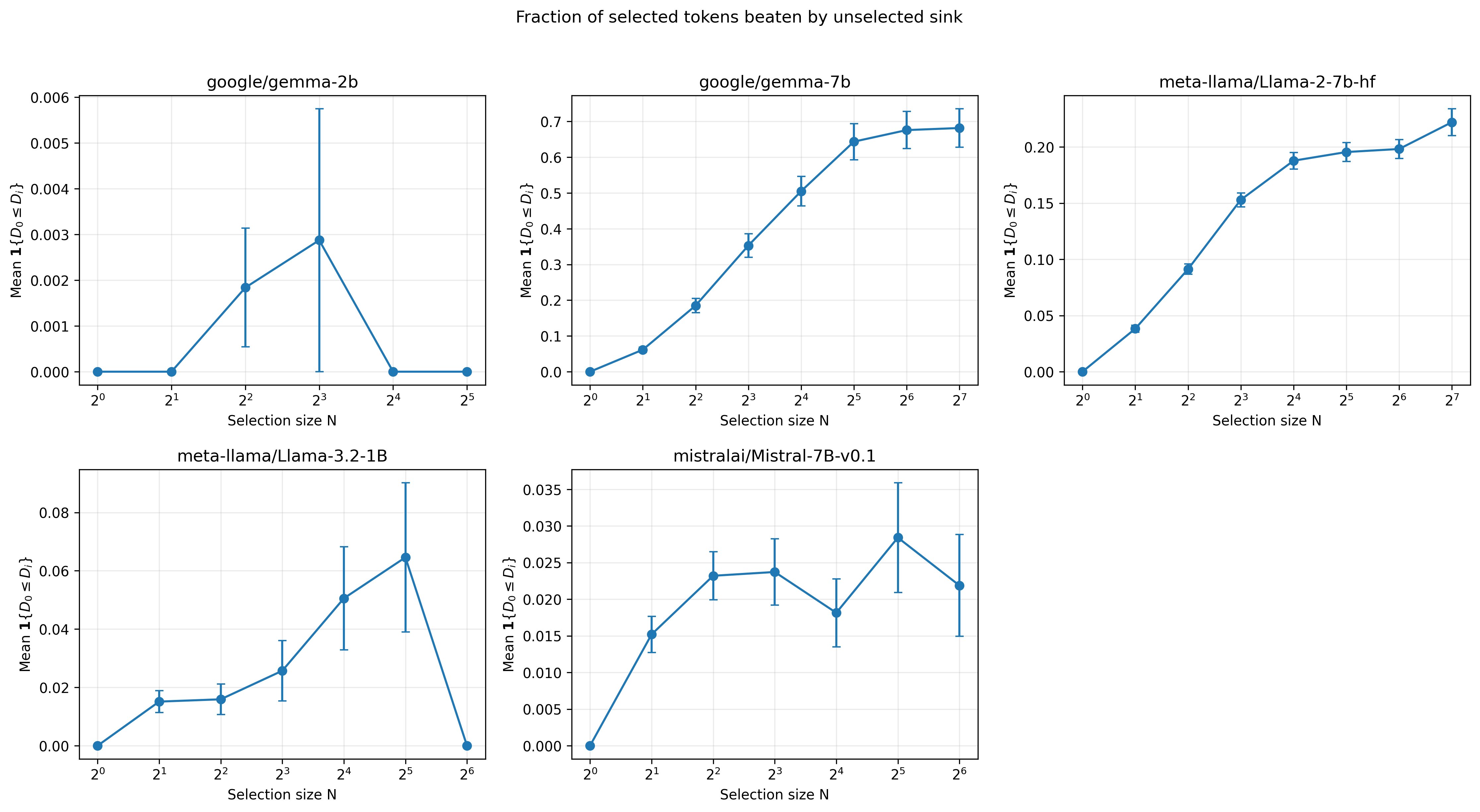}
    \caption{
    \textbf{Sink-related pairwise inversions.}
    We report the fraction of selected tokens whose distance to \(s_N\) is larger than the unselected sink distance. 
    These are precisely the sink-related  inversion events that enter the deterministic bound.
    }
    \label{fig:ap_sink_pair_inversions}
\end{figure}

Finally, we examine the alignment between the sink contribution and the selected non-sink aggregate,
\[
    \cos(y_0,s_{N,+}),
    \qquad
    s_{N,+}=\sum_{i\in I_N,\ i>0}y_i .
\]
This diagnostic explains whether the selected sink reinforces or opposes the non-sink part of the aggregate.

\begin{figure}[h]
    \centering
    \includegraphics[width=\linewidth]{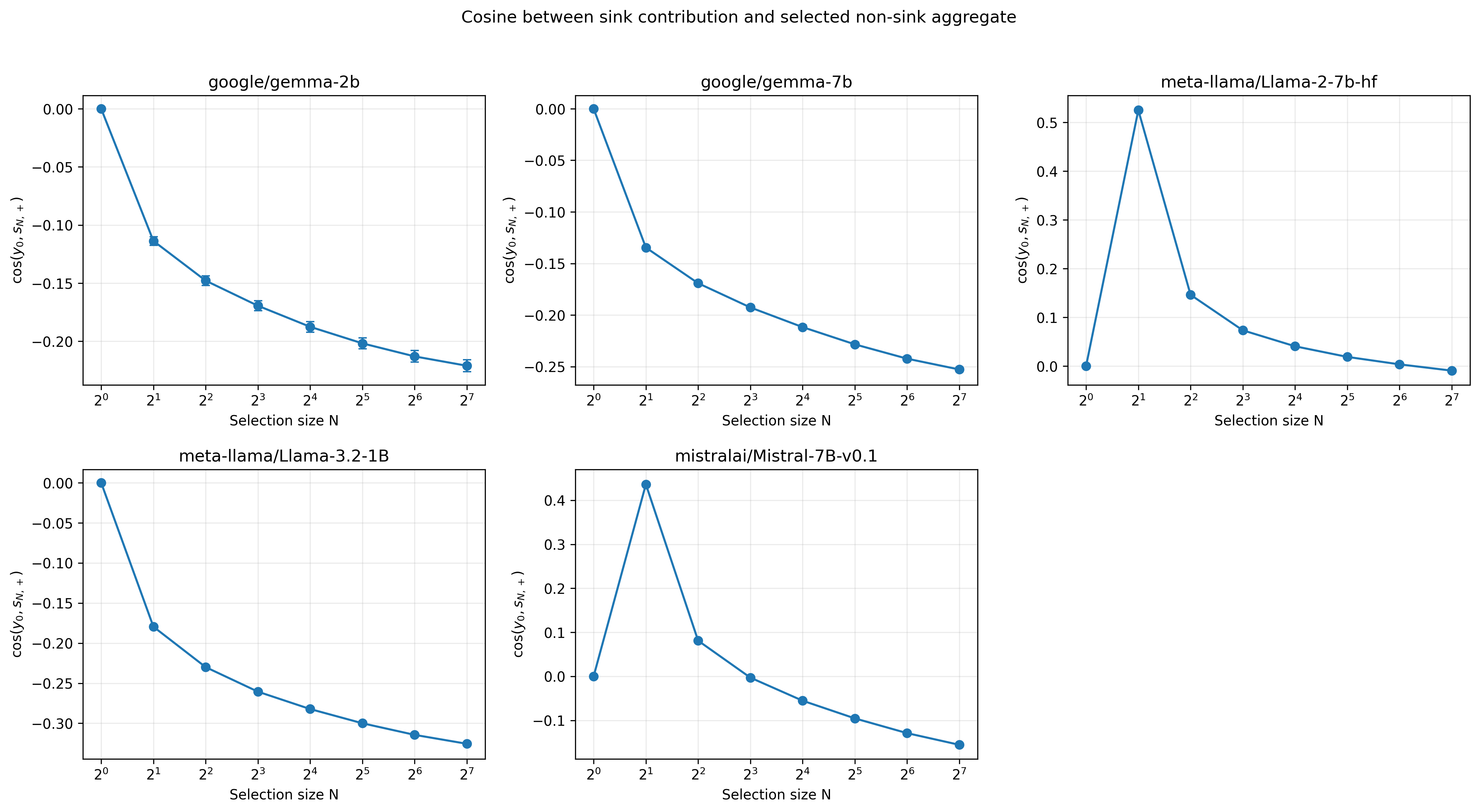}
    \caption{
    \textbf{Alignment between sink contribution and selected non-sink aggregate.}
    The sign and magnitude of \(\cos(y_0,s_{N,+})\) show whether the selected sink reinforces or opposes the non-sink selected aggregate.
    }
    \label{fig:ap_sink_selected_alignment}
\end{figure}

Overall, these diagnostics show that the sink affects geometric classification through two mechanisms. 
When selected, it can shift the aggregate and change its direction. 
When unselected, it can still intrude into the selected neighborhood and create sink-related inversions. 
This explains why the sink-aware theorem separates non-sink/non-sink pairs from sink-related pairs.

\newpage
\subsection{Head ablation sanity check}
\label{ap:ablation}

We use head ablation as a functional sanity check for the proposed geometric diagnostics. For each head, we zero out its attention output and measure the increase in next-token negative log-likelihood, denoted by $\Delta\mathrm{NLL}$. This experiment is not intended as a pruning benchmark; instead, it tests whether geometric quantities are associated with functional sensitivity.
\begin{table*}[!h]
\centering
\caption{
\textbf{Correlations between head diagnostics and ablation sensitivity.}
We report the strongest correlations with $\Delta\mathrm{NLL}$ under head ablation.
}
\label{tab:ap_ablation_correlations}
\resizebox{\linewidth}{!}{
\begin{tabular}{lllrr}
\toprule
Model & Dataset & Score & Spearman & Pearson \\
\midrule
Gemma-2B & openwebtext & \texttt{mean\_last\_attention} & 0.429 & 0.637 \\
Gemma-2B & openwebtext & \texttt{mean\_sink\_attention} & -0.386 & -0.339 \\
Gemma-2B & openwebtext & \texttt{mean\_topN\_attention\_mass} & 0.190 & 0.274 \\
Gemma-2B & openwebtext & \texttt{mean\_attention\_entropy} & -0.163 & -0.455 \\
Gemma-2B & openwebtext & \texttt{mean\_top1\_attention} & 0.103 & 0.507 \\
Gemma-2B & wikitext & \texttt{mean\_last\_attention} & 0.373 & 0.571 \\
Gemma-2B & wikitext & \texttt{mean\_sink\_attention} & -0.211 & -0.138 \\
Gemma-2B & wikitext & \texttt{mean\_attention\_entropy} & -0.208 & -0.455 \\
Gemma-2B & wikitext & \texttt{mean\_top1\_attention} & 0.206 & 0.464 \\
Gemma-2B & wikitext & \texttt{mean\_topN\_attention\_mass} & 0.195 & 0.293 \\
LLaMA-3.2-1B & openwebtext & \texttt{mean\_sink\_attention} & -0.502 & -0.323 \\
LLaMA-3.2-1B & openwebtext & \texttt{mean\_last\_attention} & 0.462 & 0.372 \\
LLaMA-3.2-1B & openwebtext & \texttt{mean\_top1\_attention} & -0.444 & -0.159 \\
LLaMA-3.2-1B & openwebtext & \texttt{mean\_attention\_entropy} & 0.267 & 0.094 \\
LLaMA-3.2-1B & openwebtext & \texttt{mean\_topN\_attention\_mass} & -0.121 & -0.064 \\
LLaMA-3.2-1B & wikitext & \texttt{mean\_last\_attention} & 0.334 & 0.343 \\
LLaMA-3.2-1B & wikitext & \texttt{mean\_sink\_attention} & -0.133 & -0.099 \\
LLaMA-3.2-1B & wikitext & \texttt{mean\_top1\_attention} & -0.097 & 0.016 \\
LLaMA-3.2-1B & wikitext & \texttt{mean\_attention\_entropy} & 0.064 & -0.044 \\
LLaMA-3.2-1B & wikitext & \texttt{mean\_topN\_attention\_mass} & -0.022 & 0.062 \\
Mistral-7B & openwebtext & \texttt{mean\_sink\_attention} & -0.312 & -0.043 \\
Mistral-7B & openwebtext & \texttt{mean\_top1\_attention} & -0.303 & -0.007 \\
Mistral-7B & openwebtext & \texttt{mean\_last\_attention} & 0.264 & 0.020 \\
Mistral-7B & openwebtext & \texttt{mean\_attention\_entropy} & 0.239 & 0.012 \\
Mistral-7B & openwebtext & \texttt{mean\_topN\_attention\_mass} & -0.146 & -0.013 \\
Mistral-7B & wikitext & \texttt{mean\_last\_attention} & 0.175 & 0.079 \\
Mistral-7B & wikitext & \texttt{mean\_sink\_attention} & -0.153 & -0.092 \\
Mistral-7B & wikitext & \texttt{mean\_top1\_attention} & -0.141 & -0.065 \\
Mistral-7B & wikitext & \texttt{mean\_attention\_entropy} & 0.098 & 0.022 \\
Mistral-7B & wikitext & \texttt{mean\_topN\_attention\_mass} & -0.037 & 0.020 \\
\bottomrule
\end{tabular}}
\end{table*}

These correlations are not meant to identify a complete head-importance metric. Instead, they show that simple geometric diagnostics have non-zero association with functional sensitivity. This supports the interpretation of the proposed geometry as a useful diagnostic signal, while leaving robust pruning as future work.

\subsection{Gemma-7B critical-head diagnostic}
\label{ap:gemma_critical_head}

The raw Gemma-7B sparsification curves are dominated by one head at layer 0,
head 12. Removing this head alone increases decode NLL by \(79.78\), whereas
the next-largest single-head increase is \(1.68\). Random masks consequently
split into two populations. At \(75\%\) retention, masks that remove the
critical head produce \(\Delta\mathrm{NLL}\) between approximately \(113\) and
\(233\), while masks that retain it are mostly between \(0.4\) and \(8.1\).
The large raw mean is therefore not a plotting artifact or a single evaluation
example; it is a mask-level critical-head event. In Figure~\ref{fig:ablation}
we hold this head fixed and exclude it from the removable ranking. The
unfiltered measurements remain available in the released result tables.

This diagnostic also motivates reporting both mean and median head-ablation
impact when ranking regimes. For Gemma-7B Reset heads, the mean is dominated by
the critical head whereas the median is close to zero, so a type mean alone is
not a robust estimate of typical head importance.

\newpage
\subsection{Taxonomy across models}
\label{ap:taxonomy}

\begin{algorithm}[t]
\caption{Parameter-free source-winner head classification}
\label{alg:head_classification}
\begin{algorithmic}[1]
\Require A collection of \(M\) causal sequences with an initial sink/BOS at
position \(0\); attention probabilities
\(A^{(m,\ell,h)}\); value vectors \(v^{(m,\ell,h)}_j\)
\Ensure One label \(c_{\ell,h}\in\{\textsc{Retriever},\textsc{Mixer},\textsc{Reset}\}\)
for every layer--head pair

\For{each layer \(\ell\) and query head \(h\)}
    \For{each sequence \(m=1,\dots,M\)}
        \For{each query position \(i=1,\dots,L-1\)}
            \State \(j^\star \gets
            \arg\max_{0\leq j\leq i}
            A^{(m,\ell,h)}_{ij}\lVert v^{(m,\ell,h)}_j\rVert_2\)
            \If{\(j^\star=i\)}
                \State \(r_{m,i}\gets\textsc{Diagonal}\)
            \ElsIf{\(j^\star=0\)}
                \State \(r_{m,i}\gets\textsc{Sink}\)
            \Else
                \State \(r_{m,i}\gets\textsc{Other}\)
            \EndIf
        \EndFor
        \State \(r_m\gets\operatorname{Plurality}_{i}(r_{m,i})\)
        \Comment{Map an exact tie to \textsc{Other}.}
    \EndFor
    \State \(r_{\ell,h}\gets\operatorname{Plurality}_{m}(r_m)\)
    \Comment{Map an exact tie to \textsc{Other}.}
    \If{\(r_{\ell,h}=\textsc{Diagonal}\)}
        \State \(c_{\ell,h}\gets\textsc{Retriever}\)
    \ElsIf{\(r_{\ell,h}=\textsc{Sink}\)}
        \State \(c_{\ell,h}\gets\textsc{Mixer}\)
    \Else
        \State \(c_{\ell,h}\gets\textsc{Reset}\)
    \EndIf
\EndFor
\State \Return \(\{c_{\ell,h}\}\)
\end{algorithmic}
\end{algorithm}

The procedure has no decision thresholds, relative margins, learned weights,
or model-specific calibration. The only deterministic tie rule maps ambiguity
to \textsc{Other}/\textsc{Reset}. We exclude query position \(0\), where the
sink and diagonal categories are identical. All counts below use the equalized
single-document preprocessing described with Table~\ref{tab:models}.

\begin{table}[!h]
\centering
\caption{\textbf{Distribution of parameter-free geometric head regimes.}
Retriever is the diagonal/current-token winner, Mixer is the sink winner, and
Reset denotes a non-privileged context winner. Counts are reported separately
for OpenWebText and WikiText after equalized preprocessing.}
\label{tab:taxonomy_distribution}
\resizebox{\linewidth}{!}{
\begin{tabular}{llrrrr}
\toprule
Model & Dataset & Total & Retriever & Mixer & Reset \\
\midrule
LLaMA-2-7B & OpenWebText & 1024 & 67 (6.5\%) & 204 (19.9\%) & 753 (73.5\%) \\
LLaMA-2-7B & WikiText & 1024 & 75 (7.3\%) & 245 (23.9\%) & 704 (68.8\%) \\
LLaMA-3.2-1B & OpenWebText & 512 & 57 (11.1\%) & 307 (60.0\%) & 148 (28.9\%) \\
LLaMA-3.2-1B & WikiText & 512 & 55 (10.7\%) & 311 (60.7\%) & 146 (28.5\%) \\
Mistral-7B & OpenWebText & 1024 & 61 (6.0\%) & 421 (41.1\%) & 542 (52.9\%) \\
Mistral-7B & WikiText & 1024 & 59 (5.8\%) & 421 (41.1\%) & 544 (53.1\%) \\
Gemma-2B & OpenWebText & 144 & 21 (14.6\%) & 51 (35.4\%) & 72 (50.0\%) \\
Gemma-2B & WikiText & 144 & 22 (15.3\%) & 46 (31.9\%) & 76 (52.8\%) \\
Gemma-7B & OpenWebText & 448 & 77 (17.2\%) & 227 (50.7\%) & 144 (32.1\%) \\
Gemma-7B & WikiText & 448 & 76 (17.0\%) & 229 (51.1\%) & 143 (31.9\%) \\
\bottomrule
\end{tabular}
}
\end{table}

\subsection{Geometrical representation}
\label{ap:geom}
\begin{figure}[h]
    \centering
    \includegraphics[width=\linewidth]{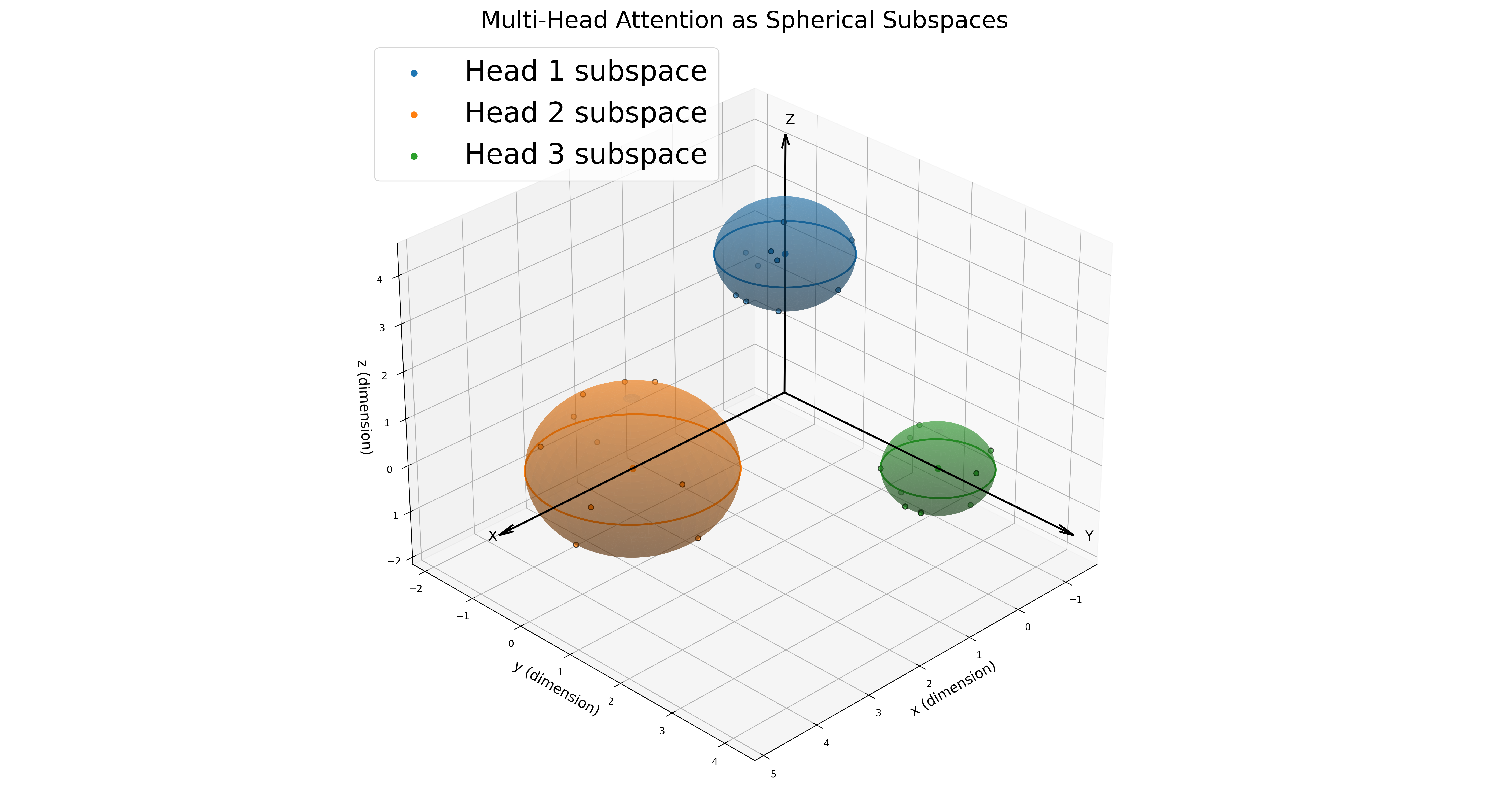}
    \caption{Representation of multi-head spherical geometry of the attention mechanism.}
    \label{fig:spherical_logic}
\end{figure}

Figure\ref{fig:spherical_logic} provides a geometric visualization of the multi-head attention mechanism, where each head is represented as a spherical subspace in three dimensions. The spheres capture how different attention heads project the input representations into distinct subspaces, thereby enabling the model to capture diverse contextual relationships. This spherical abstraction emphasizes the complementary nature of multiple heads, illustrating how they jointly span different regions of the representation space to enrich feature extraction.

\clearpage

%%%%%%%%%%%%%%%%%%%%%%%%%%%%%%%%%%%%%%%%%%%%%%%%%%%%%%%%%%%%%%%%%%%%%%%%%%%%%%%
%%%%%%%%%%%%%%%%%%%%%%%%%%%%%%%%%%%%%%%%%%%%%%%%%%%%%%%%%%%%%%%%%%%%%%%%%%%%%%%

\end{document}